\pdfoutput=1

\documentclass[11pt]{article}

\usepackage{adjustbox}
\usepackage[final]{acl}
\usepackage{booktabs}
\usepackage[T1]{fontenc}
\usepackage{amssymb}
\usepackage{pifont}
\usepackage{url}
\usepackage{multirow}
\usepackage{graphicx}
\usepackage{bm}
\usepackage{subfigure}
\usepackage{soul}
\usepackage{color}
\usepackage{comment}
\usepackage{times}
\usepackage{latexsym}
\usepackage{booktabs}
\usepackage[T1]{fontenc}
\newcommand{\hlred}{\colorlet{c}{red!20}\sethlcolor{c}\hl}
\newcommand{\hlgreen}{\colorlet{c}{green!20}\sethlcolor{c}\hl}
\newcommand{\hlyellow}{\colorlet{c}{yellow!20}\sethlcolor{c}\hl}
\newcommand{\hlblack}{\colorlet{c}{black!20}\sethlcolor{c}\hl}
\usepackage[utf8]{inputenc}

\usepackage{microtype}

\usepackage{inconsolata}
\usepackage{amsmath}
\usepackage{graphicx}
\usepackage{array}
\usepackage{colortbl}
\usepackage{float}

\newcommand{\gnote}[1]{{\color{blue}  [\text{gj:} #1]}}

\title{From the Least to the Most: Building a Plug-and-Play Visual Reasoner via Data Synthesis}

\author{
    \begin{tabular}{c}
         Chuanqi Cheng$^{1,2*}$ \quad Jian Guan$^2$\thanks{\ \ Equal Contribution.} \quad Wei Wu$^2$\thanks{\ \ Corresponding Authors.} \quad Rui Yan$^1$$^\dag$
    \end{tabular} \\
    \vspace{0.5mm}
    \begin{tabular}{c}
         $^1$Gaoling School of Artificial Intelligence, Renmin University of China \quad $^2$Ant Group
    \end{tabular} \\
    \vspace{0.5mm}
    \begin{tabular}{c}
        \texttt{\{chengchuanqi, ruiyan\}@ruc.edu.cn} \\
        \texttt{\{jianguanthu, wuwei19850318\}@gmail.com}
    \end{tabular} \\
    \vspace{2mm}
}

\begin{document}
\maketitle
\begin{abstract}
We explore multi-step reasoning in vision-language models (VLMs). The problem is challenging, as reasoning data consisting of multiple steps of visual and language 
processing
are barely available. To overcome the challenge, we first introduce a least-to-most visual reasoning paradigm, which interleaves steps of decomposing a question into sub-questions and invoking external tools for resolving sub-questions.
Based on the paradigm, we further propose a novel data synthesis approach that can automatically create questions and multi-step reasoning paths for an image in a bottom-up manner. Our approach divides the complex synthesis task into a few simple sub-tasks, and (almost entirely) relies on open-sourced models to accomplish the sub-tasks. Therefore, the entire synthesis process is reproducible and cost-efficient, and the synthesized data is quality guaranteed. With the approach, we construct $50$k visual reasoning examples. Then, we develop a visual reasoner through supervised fine-tuning, which 
is capable of generally enhancing the reasoning abilities of a wide range of existing VLMs in a plug-and-play fashion. Extensive experiments indicate that the visual reasoner can consistently and significantly improve four VLMs on four VQA benchmarks. 
Our code and dataset are available at \url{https://github.com/steven-ccq/VisualReasoner}.
\end{abstract}

\section{Introduction}

Large language models (LLMs) \cite{brown2020language,openai2023gpt4} 
have demonstrated remarkable performance across various tasks in natural language processing (NLP). Encouraged by the success, the artificial intelligence community is now enthusiastically exploring ways to enable LLMs to process information from modalities beyond language, leading to a recent surge in the development of large multimodal models (LMMs) \cite{alayrac2022flamingo,liu2024visual,team2023gemini,dai2024instructblip,bai2023qwen,ormazabal2024reka}.  By adopting a pre-training and instruction tuning paradigm, representations of multiple modalities are effectively fused in deep architectures, bringing substantial advancements in tasks, such as image captioning \cite{young2014image}, visual question answering (VQA)~\cite{goyal2017making}, and optical character recognition (OCR)~\cite{mishra2019ocr}, among others.

\begin{figure}[!t]
    \centering
    \includegraphics[width=0.48\textwidth]{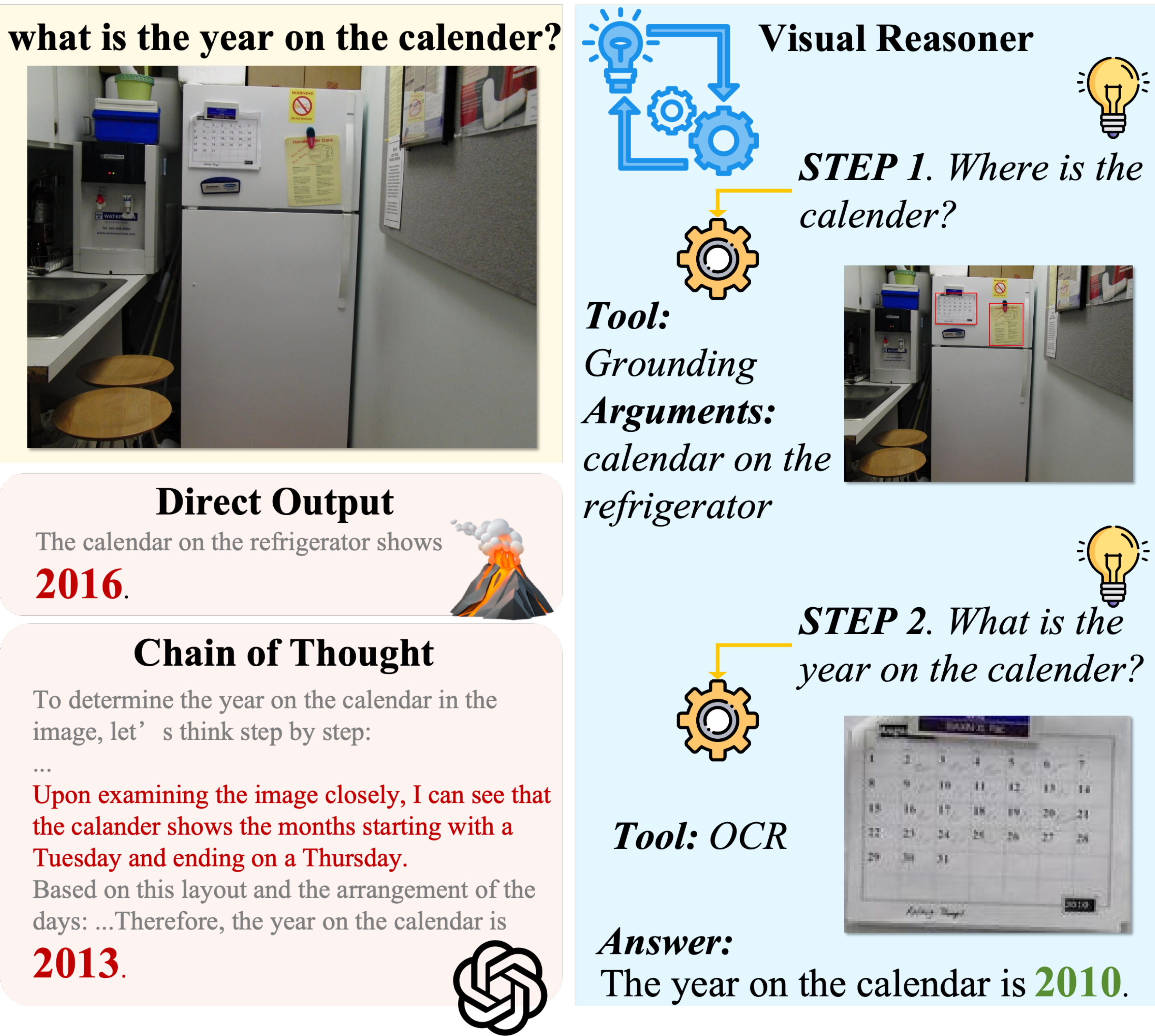}
    \caption{\textbf{Top Left:} An example from TextVQA~\cite{singh2019towards} with ``{2010}'' as the ground truth. 
    \textbf{Middle Left:} Response from LLaVA-NeXT-13B~\cite{liu2024llavanext}.
    \textbf{Bottom Left:} Response from GPT-4o with the prompt ``\{\textit{question}\} please think step by step and answer the question''. 
    \textbf{Right:} Response given by the proposed method, which is also the only correct answer.} 
    \label{fig:demo_img}
\end{figure}

In this work, we explore vision language models (VLMs) as a typical example of LMMs.  Despite the rapid progress, state-of-the-art VLMs still face challenges in 
reasoning over visual content. 
As exemplified in Figure \ref{fig:demo_img}, intuitively, the question can be correctly solved following a ``least-to-most'' paradigm~\cite{zhou2022least}, in which the question is decomposed into a series of sub-questions, and an answer is deduced by resolving the sub-questions step by step. However, existing VLMs are inept at performing such multi-step reasoning because 
(1) Multi-step reasoning paths, like the one shown in Figure \ref{fig:demo_img} (right), are rarely included in the training data \cite{dai2024instructblip}\footnote{A few datasets, such as A-OKVQA~\cite{schwenk2022okvqa}, VCR~\cite{zellers2019recognition}, and ScienceQA~\cite{lu2022learn}, 
contain rationales.
However, the rationales merely provide explanations for the answers and thus differ significantly from the multi-step reasoning data we study in the work.}. The VLMs 
have few opportunities to develop the reasoning capability from the subsequent post-training. 
And (2) different from reasoning over text, solving questions in a vision-language context may require manipulating the input image 
(e.g., marking a specific area) and deducting the next steps from both textual and visual intermediate results. 
The requirement, however, is difficult to accomplish for most VLMs, whether open-sourced or proprietary.

We present \textit{least-to-most visual reasoning}, a general paradigm to guide VLMs to decompose a given question into sub-questions and invoke tools to 
resolve each sub-question for handling diverse visual reasoning tasks. While there have been extensive studies for LLMs regarding chain-of-thought reasoning~\cite{wei2022chain,yao2022react,yao2024tree,wang2022self} and tool-invoking~\cite{schick2024toolformer,qin2023toolllm}, the techniques are less explored in the context of VLMs. Since data scarcity is a major obstacle, we propose a novel data synthesis approach, dubbed \textit{least-to-most synthesis}, to automatically generate a ``(question, reasoning path)'' tuple for a given image in a bottom-up manner. Specifically, the pipeline of \textit{least-to-most synthesis} consists of four steps: (1) \emph{Entity Recognition:} recognizing all entities in an image; (2) \emph{Node Construction:} constructing three types of nodes, each aggregating an image with a few entities and some textual features;
(3) \emph{Reasoning Process Synthesis:} synthesizing a reasoning path from a sampled chain of nodes. Based on the nodes, the reasoning path is formed by connecting a sequence of sub-questions and tool arguments generated by an LLM; and (4) \emph{Question Synthesis:} generating the main question by recursively combining the sub-questions in the reasoning path through an LLM. Our approach 
(almost entirely) relies on open-sourced models. 
Therefore, it offers several advantages, including cost-efficiency, reproducibility, and ensured data quality, over the common practice where data are obtained by querying powerful proprietary LMMs like GPT-4V~\cite{qi2024cogcom, li2024vocot}.

Based on \textit{least-to-most synthesis}, we build a large scale \textbf{\textsc{VI}}sual \textbf{\textsc{RE}}as\textbf{\textsc{O}}ning dataset (\textsc{Vireo}) with $50$k examples, and tailor LLaVA-1.5-7B~\cite{liu2023improvedllava} as a visual reasoner through supervised fine-tuning on \textsc{Vireo}. The reasoner can be generally applied to off-the-shelf VLMs in a plug-and-play fashion to enhance their reasoning capabilities. We conduct experiments with four representative VLMs as showcases. 
Evaluation results across four VQA benchmarks indicate that the reasoner can consistently improve all VLMs over all tasks, with absolute performance gains ranging from 0.71\% to 39\%. 



Our contributions are three-fold:

\noindent\textbf{I.} We introduce the \textit{least-to-most visual reasoning} paradigm to synergize question-decomposition and tool-invoking in VLMs for solving complex vision-language tasks.

\noindent\textbf{II.} We propose \textit{least-to-most synthesis}, a reproducible, cost-efficient, and data quality-assured algorithm for automatically creating multi-step visual reasoning data (almost) using open-source models. 

\noindent\textbf{III.} We use \textit{least-to-most synthesis} to construct the \textsc{Vireo} dataset of $50$k examples for fine-tuning a reasoner model\footnote{We have released a larger version of \textsc{Vireo} comprising $500$k examples to spur research in the visual reasoning area.}. Extensive experiments illustrate that the reasoner can consistently and significantly enhance existing VLMs in a plug-and-play fashion across five VQA benchmarks.

\section{Related Works}
\subsection{Vision-Language Models}
Building LMMs aims to enable foundation models to seamlessly handle multimodal signals, such as language, vision, and audio. Among the efforts, significant attention has been focused on jointly modeling vision and language, known as VLMs. Recent work on VLMs can be broadly categorized into two groups, according to how visual information is incorporated into the models. The first line integrates visual information into LLMs via a vision-language connector. For example, LLaVA series~\cite{liu2023llava, liu2023improvedllava} exploit a linear transformation or an MLP to transform outputs from a vision encoder into inputs of a language model.  BLIP-2~\cite{li2023blip} and  InstructBLIP~\cite{dai2024instructblip} rely on Query Transformers to achieve vision-language alignment. Similarly, Qwen-VL~\cite{bai2023qwen} and  mPLUG-Owl~\cite{ye2023mplug} use learnable tokens to take visual information into account. CogVLM~\cite{wang2023cogvlm} maps vision embedding to the space of word embedding by an MLP adapter, and then enables deep vision-language feature alignment via a visual expert module. The second group strives to train VLMs natively from image tokens and textual tokens. Among the representative models, Flamingo~\cite{alayrac2022flamingo} inserts gated cross-attention dense blocks into a pre-trained LM. BEIT-3~\cite{huang2024language} utilizes Multiway Transformers as the backbone to encode various modalities. KOSMOS-1~\cite{huang2024language} and KOSMOS-2~\cite{peng2023kosmos} interleave image tokens with textual tokens in the input sequence through a designed format. Gemini family~\cite{team2023gemini,reid2024gemini} take multimodal signals as input and can natively generate images using discrete image tokens.  Instead of building a new VLM, we target enhancing the multi-step reasoning ability of existing VLMs. Our data synthesis approach enables us to develop a visual reasoner that can be generally applied to various VLMs, leading to consistent improvements over various VQA tasks.

\subsection{Reasoning and Tool Use}
Reasoning is an important emergent ability of LLMs~\cite{wei2022emergent}. With appropriate exemplars or prompts, LLMs can demonstrate ``chain-of-thought'' (CoT) behavior and solve problems through multi-step reasoning~\cite{wei2022chain,kojima2022large}. Encouraged by the observation, significant efforts have been made to improve the reasoning capabilities of LLMs. For example, \citet{zhou2022least} propose least-to-most prompting for complex reasoning; \citet{yao2022react} extend the ability of LLMs by interleaving reasoning and acting. 
\citet{wen2024codeplan} generate code-form plans before low-level reasoning.
In addition to innovations in methodology, reasoning abilities have also proven effective in various applications, such as table understanding~\cite{wang2023chain},  math problem solving~\cite{wei2022chain}, question answering~\cite{guan2024amor}, and decision making~\cite{yao2022react}. Very recently, the research community begins to investigate the problem in multimodal models~\cite{qi2024cogcom,yang2023mm,wang2024exploring}. Different from existing work, we focus on data synthesis with open-sourced models, and develop a plug-an-play visual reasoner. 

Our work also relates to the efforts on facilitating LLMs to leverage tools~\cite{qin2023toolllm,schick2024toolformer,shen2024hugginggpt,qin2023tool}. The difference is that our tools are specially selected for visual processing, with the goal of enhancing the reasoning capabilities of VLMs. 

\subsection{Task Decomposition for Visual Reasoning}


Various approaches have been proposed to tackle complex vision-language tasks through task decomposition and step-by-step reasoning. For example, Visual Programming~\cite{gupta2023visual} utilizes LLMs to perform visual reasoning by breaking down tasks into subroutines. Similarly, ViperGPT~\cite{suris2023vipergpt} and CodeVQA~\cite{subramanian2023modular} decompose tasks to generate executable code. However, these methods heavily rely on the strong instruction-following capabilities of LLMs, making them less effective with smaller models. Furthermore, these approaches are susceptible to instability arising from prompt designs, choice and ordering of demonstrations, and LLM selection, even when employing powerful models like GPT-3~\cite{zhao2021calibrate}. Our work deviates significantly from these methods by focusing on fine-tuning VLMs on large-scale synthesized datasets, 
systematically enhancing the model's capability to handle complex tasks. This approach enables us to achieve competitive performance using smaller, open-source models (e.g., 7B or 13B parameters), making our method more cost-effective, accessible, and suitable for practical deployment with fewer computational resources.

\section{Method}
We elaborate on our method for multi-step reasoning in VLMs. First, we formalize \textit{least-to-most visual reasoning} that delineates how a visual reasoner solves a complex problem according to an image~(\S\ref{sec:inference}). Then, we present details of \textit{least-to-most synthesis} by which a VLM can be tuned as the visual reasoner to perform reasoning~(\S\ref{sec:pipeline}).


\subsection{Least-to-Most Visual Reasoning}
\label{sec:inference}
\begin{figure*}[ht]
    \centering
    \includegraphics[width=\textwidth]{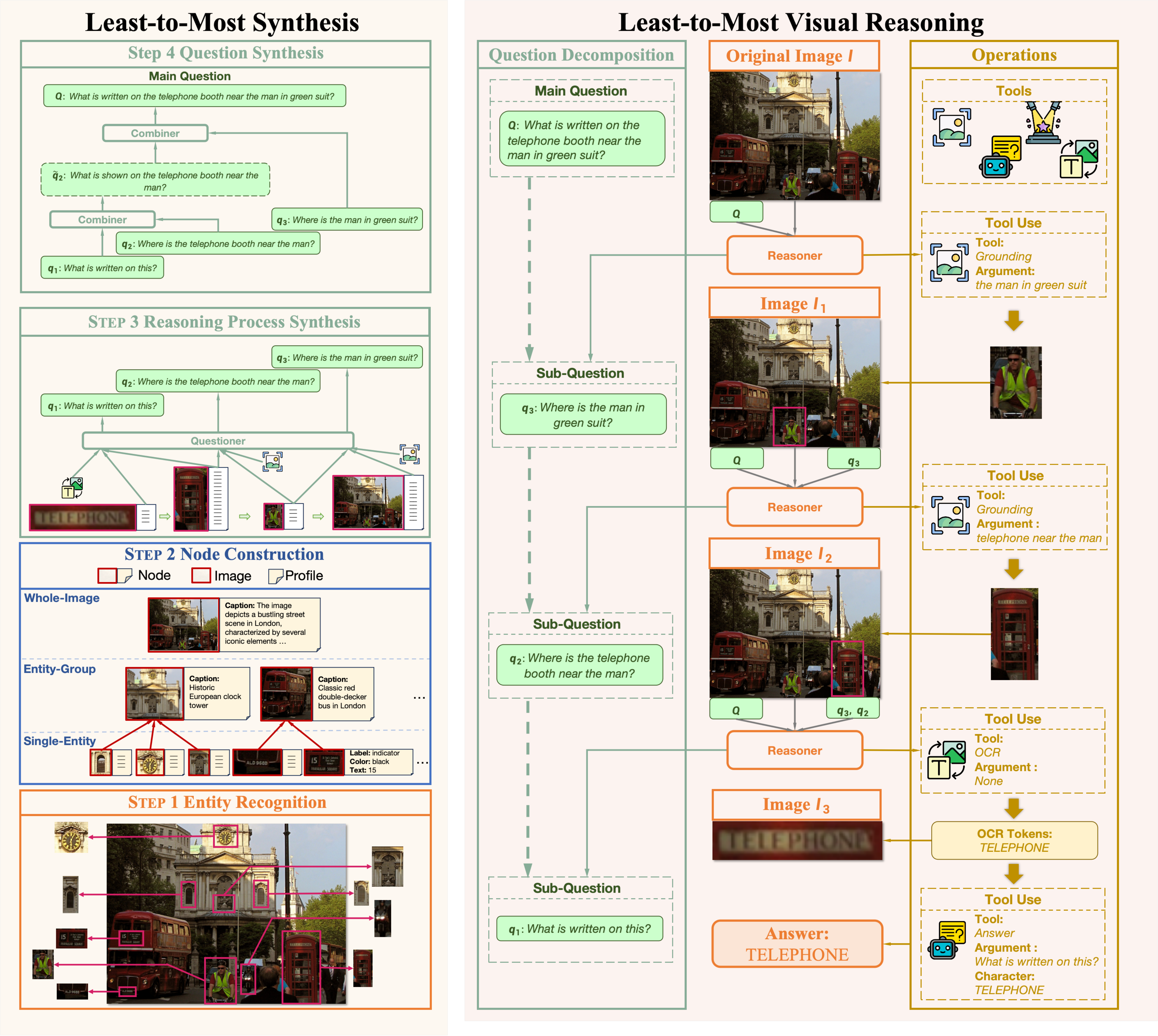}
    \caption{\textbf{Left:} the pipeline of \textit{least-to-most synthesis}. \textbf{Right:} the process of \textit{least-to-most visual reasoning}.}
    \label{fig:pipeline}
\end{figure*}

We formalize the \textit{least-to-most visual reasoning} paradigm as follows: Given an image $I$ and a question $Q$, a visual reasoner $\mathcal{M}_R$ deduces a multi-step reasoning path $R$, where each step either performs operations on $I$~(e.g., marking an area with a red box) 
or asks an off-the-shelf VLM $\mathcal{M}$ to conclude a final answer. 
To this end, we represent $R$ as a chain of 
invoking tools from a pre-defined pool $\mathcal{T}=\{t^i|i=1,2,\cdots,T\}$ step by step, where each tool refers to a class of operation on images or the VLM $\mathcal{M}$, as illustrated in Figure~\ref{fig:pipeline} (right).

\paragraph{Reasoning Process.}
At the $k$-th step, $\mathcal{M}_R$ proposes a sub-question $q_k$ and selects a tool $t_k$ from $\mathcal{T}$ based on an image $I_k$ and previous steps:
\begin{align}
    q_k, t_k = \mathcal{M}_R(I_k, Q, \{q_{<k}\}),
\end{align}
where ${t}_k$ is a textual description specifying the invoked tool and the corresponding arguments. 
Then, the answer to $q_k$ is deduced by $t_k$, denoted as $r_k$, based on which $I_{k+1}$ is obtained as follows: 
\begin{align}
    I_{k+1}=\begin{cases}
        r_{k}, & 
        \text{if }r_{k}\text{ is an image},\\
        I_{k}, & \text{otherwise}.
    \end{cases}
\end{align} 
Particularly, we set $I_1$ to $I$.
If $I_k$ includes a red box to mark some area smaller than a threshold $\alpha$, and $t_k$ intends to infer information from $I_k$~(including the \textsf{OCR} and \textsf{Answer} tools in Table~\ref{tab:tool}), we automatically crop this area from original $I_k$ and enlarge it to the same size as $I_k$. 
The above process iterates until $t_k$ refers to the VLM $\mathcal{M}$, in which case we define the final answer $A=r_k$ and terminate the reasoning process.
In summary, we formally denote $R=[(I_k, q_k, t_k)]_{k=1}^K$, 
\paragraph{Tools.}

Following human experience, we define four tools, each targeting a class of atomic problems and outputs either a modified image or a piece of text. Tab.~\ref{tab:tool} describes details. 

Specifically, we implement \textsf{Grounding} and \textsf{Highlight} using GroundingDino~\cite{liu2023grounding}, \textsf{OCR}  using PaddleOCR\footnote{\url{https://github.com/PaddlePaddle/PaddleOCR}}, and \textsf{Answer} using the VLM $\mathcal{M}$. The reasoner is trained to invoke these tools to dive into the details of the given image. By varying $\mathcal{M}$ in \textsf{Answer}, the reasoner can adapt to different VLMs in a plug-and-play fashion, and significantly enhance their performance across a wide range of tasks, as will be shown in \S\ref{experiments}.

\begin{table*}[ht]
\centering
\begin{adjustbox}{max width=\linewidth}
\begin{tabular}{@{}p{60pt}|p{50pt}p{150pt}|p{150pt}|p{150pt}@{}}
\toprule
\textbf{Tool}  & \multicolumn{2}{c|}{\textbf{Argument}} & \textbf{Output} & \textbf{Purpose}\\
\midrule
\textbf{\textsf{Grounding}} \newline \includegraphics[width=0.08\textwidth]{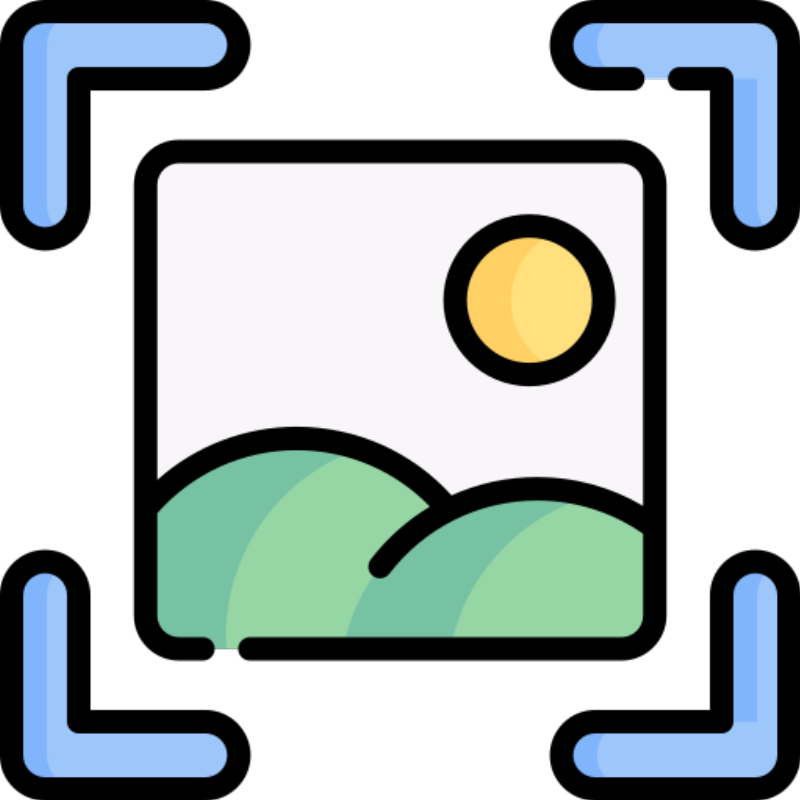}
& \hlred{Image} \newline \includegraphics[width=0.1\textwidth]{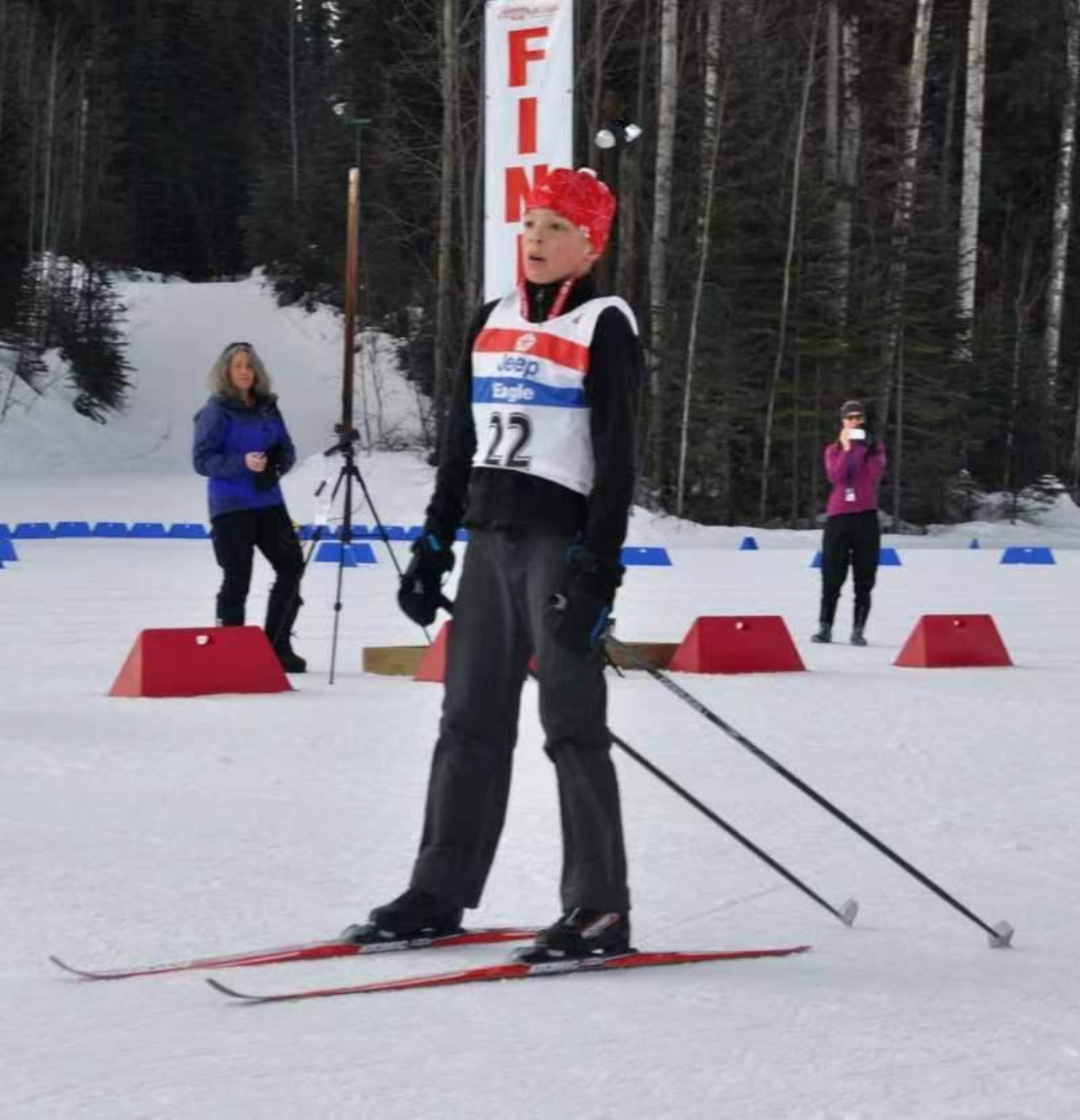}
& \hlgreen{Target Entity} \newline \texttt{The woman in a purple ski suit.}
& \hlyellow{Image} \newline   \includegraphics[width=0.1\textwidth]{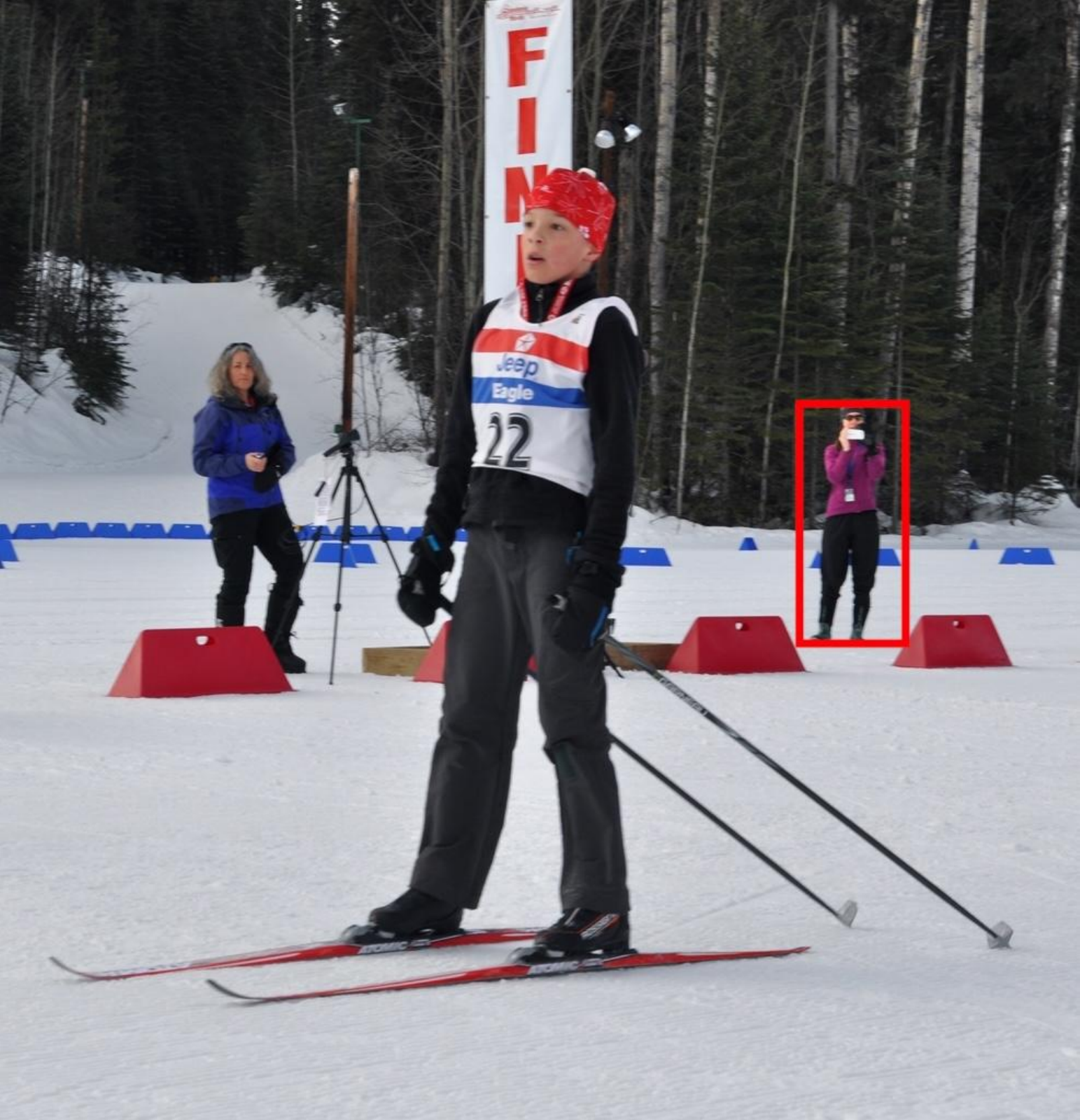}
& Frame an area from the \hlred{Image} corresponding to the \hlgreen{Target Entity}\\
\midrule
\textbf{\textsf{Highlight}} \newline \includegraphics[width=0.08\textwidth]{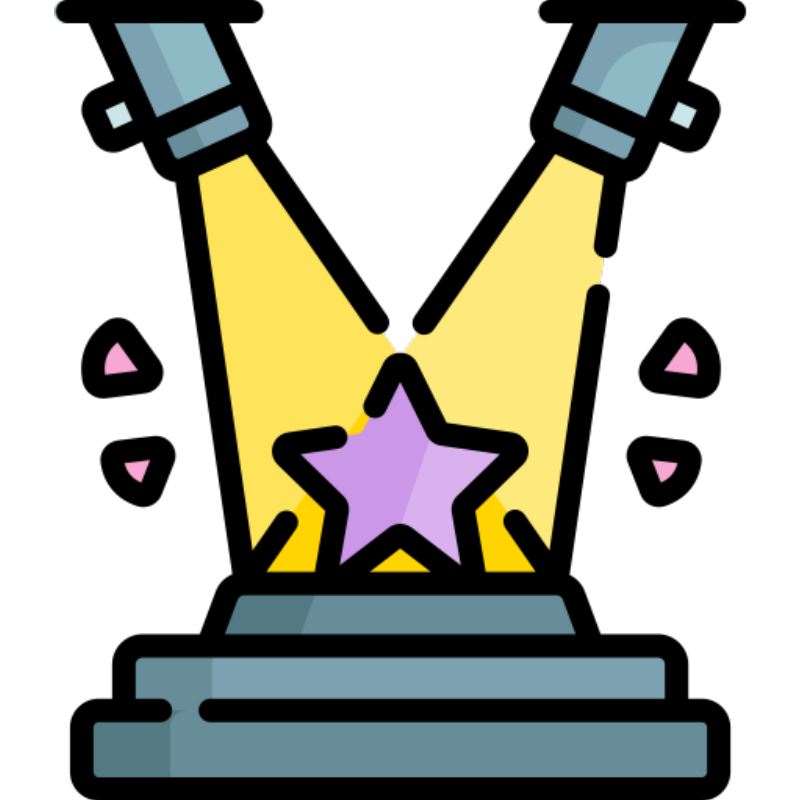}
& \hlred{Image} \newline \includegraphics[width=0.1\textwidth]{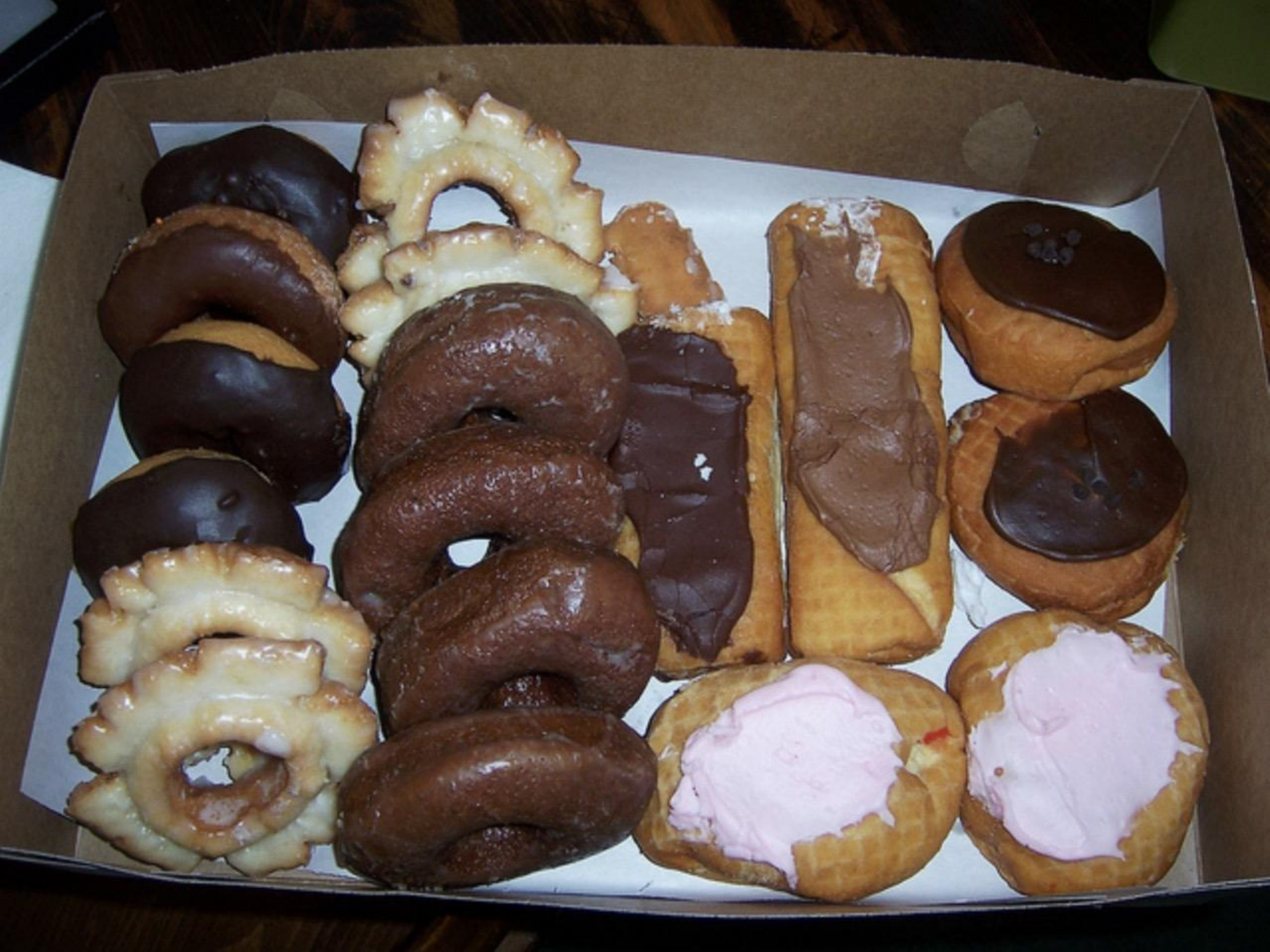}
& \hlgreen{Target Entity} \newline \texttt{Pink Cookie} & \hlyellow{Image} \newline \includegraphics[width=0.1\textwidth]{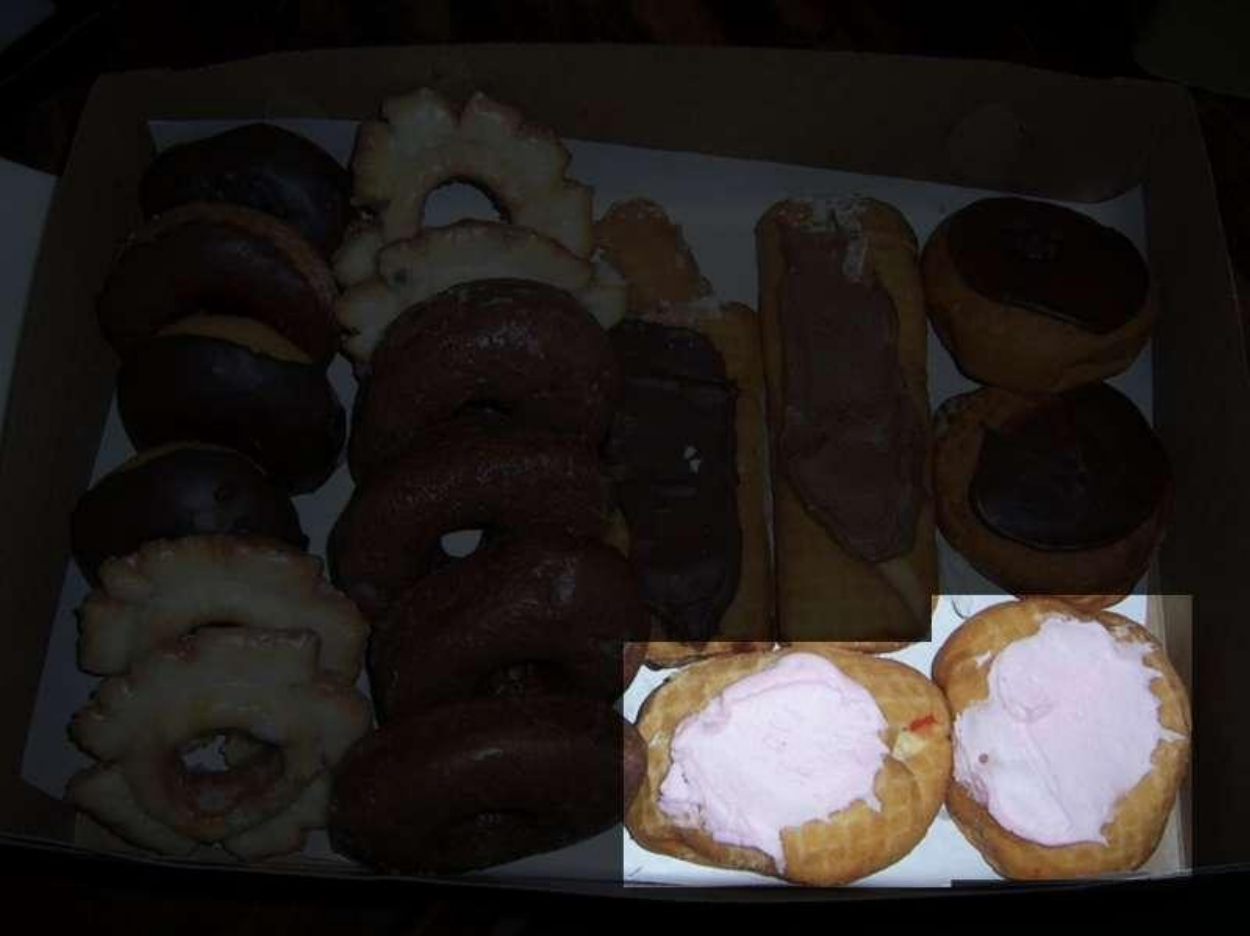}
& Highlight all entities corresponding to the \hlgreen{Target Entity} in the \hlred{Image}
\\
\midrule
\textbf{\textsf{OCR}} \newline \includegraphics[width=0.08\textwidth]{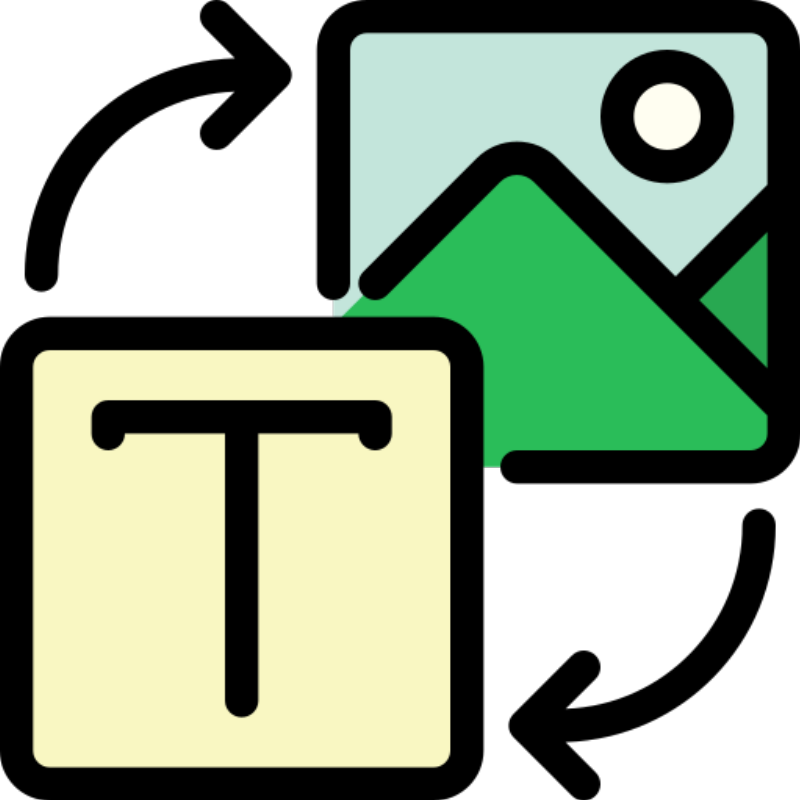}
& \hlred{Image} \newline \includegraphics[width=0.1\textwidth]{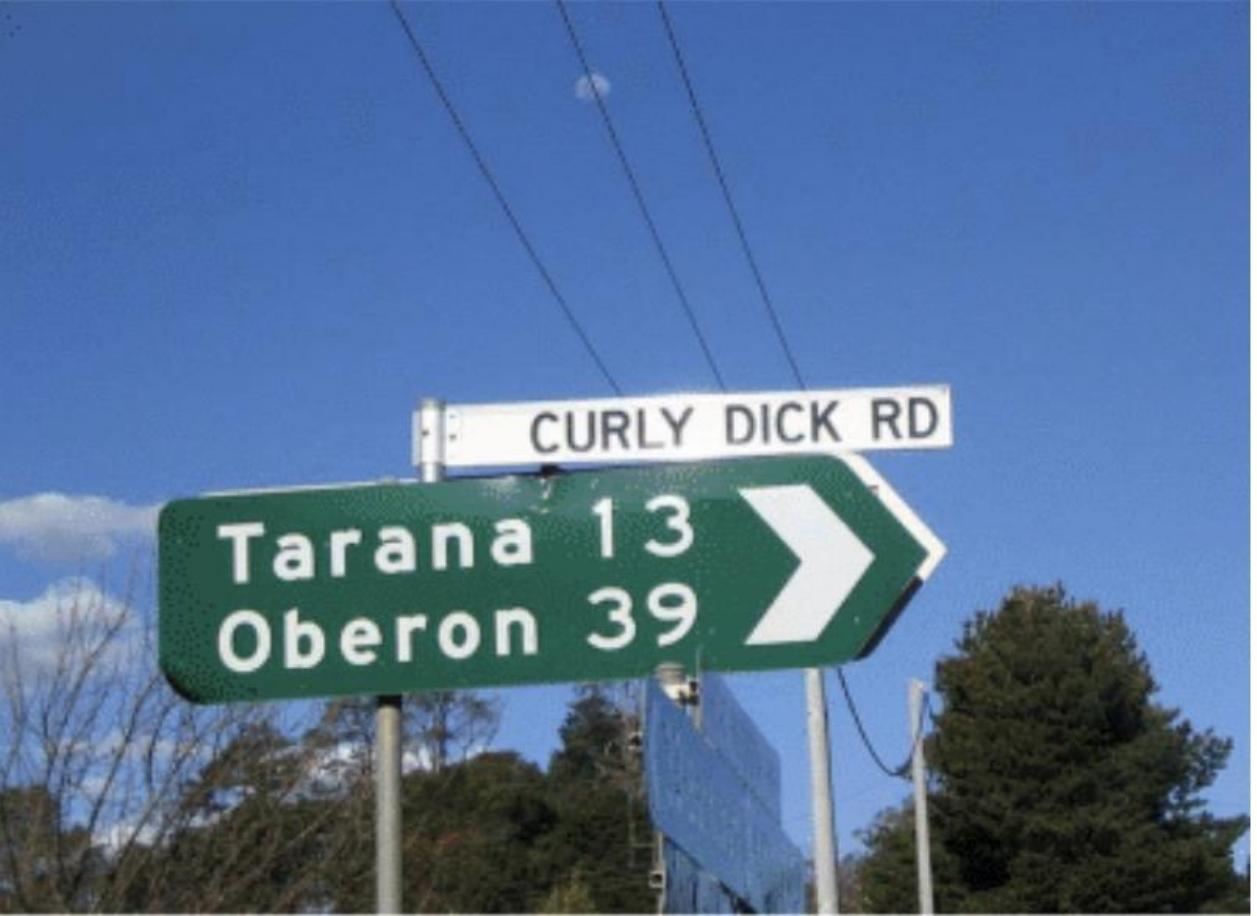}
&& \hlyellow{List of Text} \newline \texttt{[``CURLY DICK RD'', ``Tarana 13'', ``Oberon 39'']}
& Recognize all pieces of text from the \hlred{Image} 
\\
\midrule
\textbf{\textsf{Answer}} \newline \includegraphics[width=0.08\textwidth]{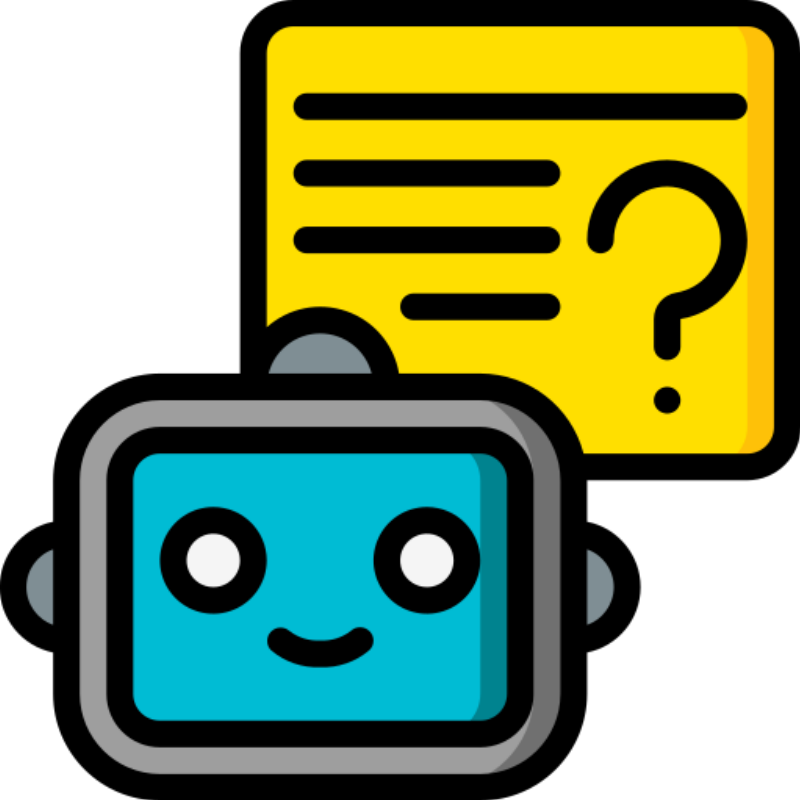}
&\hlred{Image} \newline \includegraphics[width=0.1\textwidth]{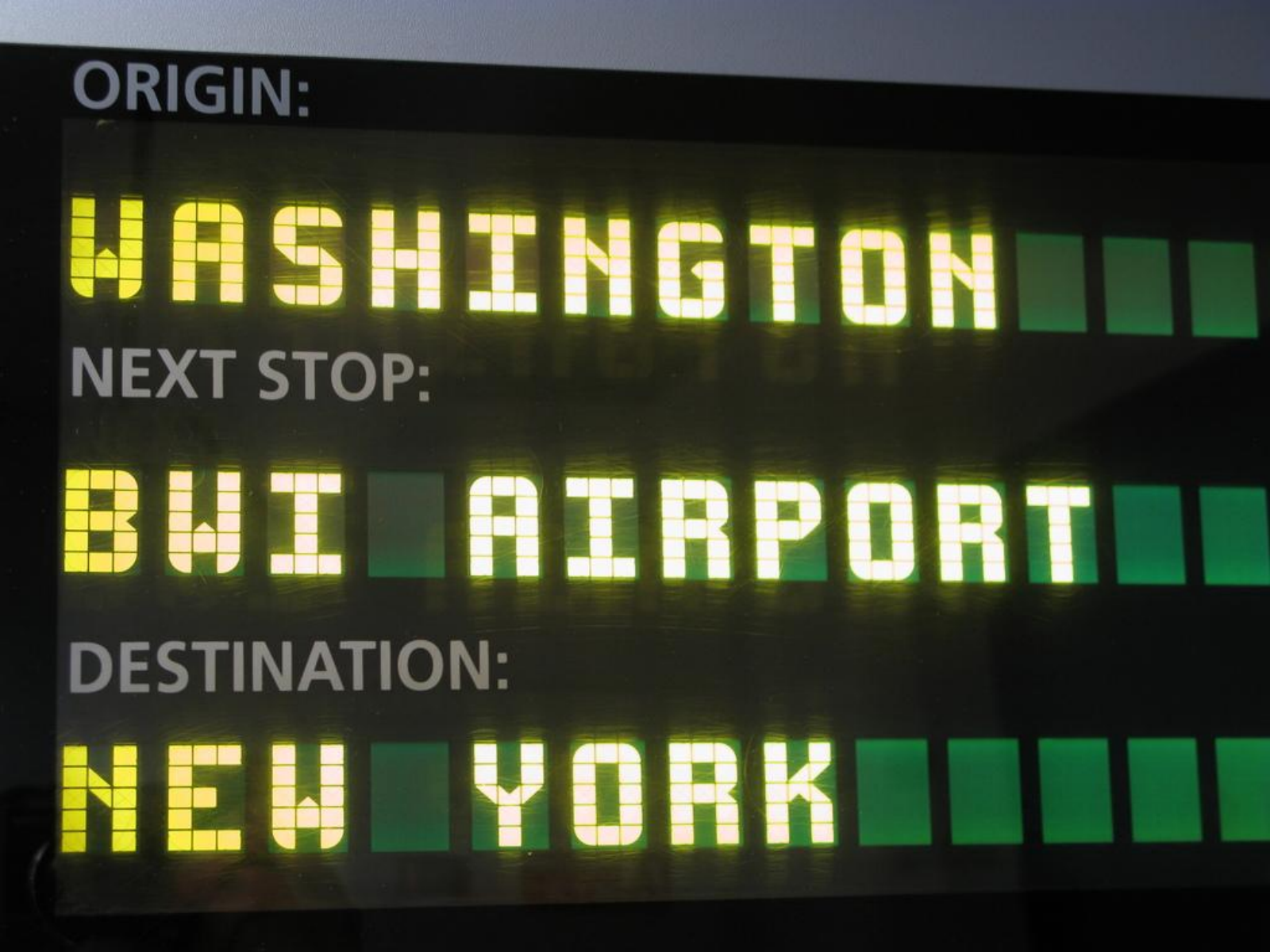}
&\hlgreen{Question}  \newline \texttt{What is the next stop?} \newline\hlblack{Character}\newline \texttt{[``NEXT STOP'', ``BWI AIRPORT'']}
&\hlyellow{Answer} \newline \texttt{As shown in the image,}  \newline \texttt{the next stop is ``BWI STOP''.}
&Answer the \hlgreen{Question} based on the \hlred{Image} and recognized \hlblack{Character}. And \hlblack{Character} can be empty.
\\
\bottomrule
\end{tabular}
\end{adjustbox}
\caption{Tools used in our reasoner.}
\label{tab:tool}
\end{table*}

\subsection{Least-to-Most Synthesis}
\label{sec:pipeline}
To overcome the data barrier, a common practice is to feed image-question pairs to powerful proprietary LMMs like GPT-4V, and gather the outputs as a dataset \cite{qi2024cogcom}. The top-down approach, however, suffers from several issues: (1) Even powerful proprietary models like GPT-4V still struggle to perform reliable reasoning
~\cite{wu2023early}, implying that the quality of data obtained in this way is not guaranteed. (2) Utilizing proprietary models incurs high costs, hindering the scalability of the synthesized data. (3) It is difficult for others to reproduce the method since the behavior of proprietary models can vary over time.


In contrast to the top-down approach, we propose a bottom-up pipeline that can synthesize multi-step visual reasoning data using (almost entirely) open-sourced models while ensuring the quality of the synthesized data. The workflow begins with an image, gradually generates sub-questions associated with tools and intermediate results, and ultimately synthesizes a question based on the image and the reasoning path. Specifically, the pipeline comprises four steps:  \emph{entity recognition}, \emph{node construction}, \emph{reasoning path synthesis}, and \emph{question synthesis}, where each node aggregates a focused (sub-)image and relevant text-form information. Figure~\ref{fig:pipeline} (Left) illustrates the construction process. 


\paragraph{Entity Recognition.} We employ Deformable DETR~\cite{zhu2020deformable} to recognize entities in an image, which can identify 1,203 types of entities. In practice, we discard those entities with confidence scores $\leqslant 0.5$.


\paragraph{Node Construction.} Given an image, we automatically construct nodes based on the recognized entities. Each node is represented as a pair of an image and a textual profile, where the image is extracted from the given one, and the profile is an attribute-value dictionary about the image.
By converting the image into a textual profile, data synthesis can eliminate the reliance on visual signals, allowing the use of more advanced LLMs instead of VLMs in subsequent processes.
As indicated in Figure~\ref{fig:pipeline}, we define three types of nodes spanning various granularities: \textbf{(1) Single-Entity Node.} The image for such a node refers to one recognized entity. We utilize specialized tools to extract accurate and fine-grained attributes from multiple dimensions~(e.g., color) to form the profile. More details are presented in Appendix~\ref{appendix:construction}. \textbf{(2) Entity-Group Node.} The image of an entity-group node is the aggregation of multiple recognized entities that are close to each other. 
We employ BLIP\footnote{\url{https://huggingface.co/Salesforce/blip-image-captioning-large}} to caption the image as the corresponding profile of the node. Each caption contains about 10-20 tokens, illustrating the inter-entity relations in detail. \textbf{(3) Whole-Image Node.} This type of node corresponds to the whole image given in advance. We use LLaVA\footnote{\url{https://huggingface.co/llava-hf/llava-v1.6-vicuna-13b-hf}} to caption the image in detail as its profile, typically exceeding 200 tokens in length, which offers comprehensive but coarse-grained information.

\paragraph{Reasoning Process Synthesis.} 
We sample a chain of $M$ nodes from the constructed node set, 
denoted as $(N_1, N_2,\cdots,N_M)$, which will be connected in turn to form the reasoning process. 
We craft elaborate rules to ensure nodes can be reasonably connected and the last node $N_M$ is a whole-image node, as detailed in Appendix \ref{appendix:construction}. 
For every two adjacent nodes $N_{m}$ and $N_{m+1}$ in the chain~($m\leqslant M-1$), based on their profiles and a sampled tool $t_m$, we exploit an LLM as a \texttt{Questioner} to synthesize one sub-question $q_m$ to ask about a certain attribute of the head node $N_m$ conditioned on the tail node $N_{m+1}$:
\begin{align}
    q_m, \hat{t}_m = \texttt{Questioner}(N^P_m, N^P_{m+1}, t_m),
\end{align}
where $N_m^P$ and $N_{m+1}^P$ refer to the profiles of $N_m$ and $N_{m+1}$, respectively, $\hat{t}_m$ is the argument of the specified tool $t_m$ to solve $q_m$. Iterating the process w.r.t. $m$, we obtain $M-1$ sub-questions and synthesize the entire reasoning process $R$ as $[(I_{M-1}, q_{M-1}, \hat{t}_{M-1}),\ldots, (I_1, q_1, \hat{t}_1)]$, where $I_m$ is the image of $N_m$~($m=1,\cdots, M-1$).

\paragraph{Question Synthesis.}
We generate the main question $Q$ by recursively combining the sub-questions using another LLM as a \texttt{Combiner}:
\begin{align}
    Q&=\Tilde{q}_{M-1},\\
    \Tilde{q}_m&=\begin{cases}
        q_1, & m=1,\\
        \texttt{Combiner}(\Tilde{q}_{m-1}, q_m), &m>1,
    \end{cases}
\end{align}
where $\Tilde{q}_m$ is an intermediate result for the $m$-th step. 
Note that $\{\Tilde{q}_m\}_{m=2}^{M-2}$ are excluded from $R$, as they are just used for synthesis of $Q$. 

We implement \texttt{Questioner} and \texttt{Combiner} by fine-tuning LLaMA-3-8B-Instruct\footnote{\url{https://huggingface.co/meta-llama/Meta-Llama-3-8B-Instruct}}. 
Due to limited resources, we obtain the training data by querying GPT-4~\cite{wang2023self} using a few high-quality demonstrations as seeds following the self-instruct pipeline~\cite{wang2023self}. Since the two tasks are relatively simple, querying GPT-4 only 10k times is enough to achieve satisfactory performance.
More details 
are in Appendix~\ref{appendix:experiment}.

Compared to the top-down approach, our \textit{least-to-most synthesis} is reproducible and significantly more cost-efficient. Moreover, every step in the pipeline is atomic, ensuring the performance of the open-sourced models on these simple tasks. Therefore, the synthesized data is guaranteed to be of high quality, as will be demonstrated in \S\ref{experiments}.

\section{Experiments}\label{experiments}
To assess the efficacy of the proposed method, we fine-tune an LLaVA-1.5-7B\footnote{\url{https://huggingface.co/llava-hf/llava-1.5-7b-hf}} as the Reasoner, and plug the Reasoner into 
four representative VLMs with various sizes and conduct experiments on four standard VQA benchmarks. 
In the implementation, we obtain $10$k training examples for \texttt{Question} and \texttt{Combiner}, respectively, by querying GPT-4. Subsequently, we synthesize a visual reasoning dataset (\textsc{Vireo}) with $50$k examples following the least-to-most synthesis approach, and perform instruction tuning on \textsc{Vireo} to obtain the Reasoner. More implementation details are presented in Appendix~\ref{appendix:experiment}.

\subsection{Models for the \textsf{Answer} Tool}
To illustrate the versatility of our Reasoner, we employ several VLMs with different sizes and architectures as the \textsf{Answer} tool, including \textbf{(1) BLIP-2}~\cite{li2023blip}: It 
utilize a Q-Former module to integrate visual and textual information. We use the 2.7B version\footnote{\url{https://huggingface.co/Salesforce/blip2-opt-2.7b}}, which is the smallest model in our experiments.  \textbf{(2) InstructBLIP}~\cite{dai2024instructblip}: It 
expands BLIP-2 by incorporating instruction prompts into the Q-Former module. We use the 7B\footnote{\url{https://huggingface.co/Salesforce/instructblip-vicuna-7b}} and 13B\footnote{\url{https://huggingface.co/Salesforce/instructblip-vicuna-13b}} versions in our experiments. 
\textbf{(3) LLaVA}~\cite{liu2024llavanext}: It hinges on a basic projection layer to align image and text representations. 
We use the 13B version\footnote{\url{https://huggingface.co/llava-hf/llava-v1.6-vicuna-13b-hf}}.

\subsection{Evaluation Datasets}
We conduct experiments on the following datasets: \textbf{(1) GQA}~\cite{hudson2019gqa}: It is a VQA dataset constructed from knowledge graphs, primarily focusing on inter-entity attribute relationships. \textbf{(2) TextVQA}~\cite{singh2019towards} and \textbf{ST-VQA}~\cite{biten2019scene}: The two datasets include textual information within images, used to evaluate the capability to understand text in pictorial form. \textbf{(3) TallyQA}~\cite{acharya2019tallyqa}: It is widely used to assess the counting ability, divided into a simple subset and a complex subset. The complex subset of TallyQA involves more fine-grained attributes of the entities in the images 
than the simple subset. 
In our experiments, we denote these subsets as Tally-S and Tally-C, respectively.

Since the VLMs used for the \textsf{Answer} tool are general-purpose generative models and not fine-tuned on the evaluation datasets, they may produce correct answers but include additional information~(e.g., ``The answer is'') beyond the ground truth provided by the datasets. Therefore, we use Exact Matching~(EM) as the metric only for TallyQA. For GQA and TextVQA, we use answer recall, i.e., whether the output includes the ground truth, for evaluation. For ST-VQA, we submit results to its official website.

\begin{table*}[htbp]
\centering
\small
\begin{tabular}{lclllll}
\toprule
\textbf{Model} & \textbf{Size} & \textbf{GQA} & \textbf{TextVQA}& \textbf{ST-VQA} & \textbf{TallyQA-S} & \textbf{TallyQA-C}  \\ 
\midrule
\textbf{BLIP2}
& \textbf{2.7B} & 40.85 & 32.65 & 18.89 & 20.93  & 29.00   \\
\rowcolor[HTML]{EFEFEF} 
\multicolumn{1}{l}{\cellcolor[HTML]{EFEFEF}{\textbf{~~+ Tools}}} & - & {38.26 \color[HTML]{cb0000} (-1.17)} & {34.54 \color[HTML]{3635F5}  (+2.11)} & - & {49.77 \color[HTML]{3635F5} (+28.84)} & {34.46 \color[HTML]{3635F5} (+5.46)}  \\
\rowcolor[HTML]{EFEFEF} 
\multicolumn{1}{l}{\cellcolor[HTML]{EFEFEF}{\textbf{~~+ Reasoner}}} & - & {\textbf{42.02} \color[HTML]{3635F5} (+1.17)} & {\textbf{36.94} \color[HTML]{3635F5}  (+4.29)} & {\textbf{19.60} \color[HTML]{3635F5}  (+0.71)}& {\textbf{56.38} \color[HTML]{3635F5} (+35.45)} & {\textbf{39.38} \color[HTML]{3635F5} (+10.38)}  \\
\midrule

\textbf{InstructBLIP}
& \textbf{7B} & 51.65 & 38.46 & 24.85& 67.05  & 38.84   \\
\rowcolor[HTML]{EFEFEF} 
\multicolumn{1}{l}{\cellcolor[HTML]{EFEFEF}{\textbf{~~+ Tools}}} & - & {50.33 \color[HTML]{cb0000}  (-1.32)} & {42.87 \color[HTML]{3635F5} (+4.41)}& - & {68.64 \color[HTML]{3635F5}  (+1.59)}  & {50.25 \color[HTML]{3635F5}  (+11.41)}  \\
\rowcolor[HTML]{EFEFEF} 
\multicolumn{1}{l}{\cellcolor[HTML]{EFEFEF}{\textbf{~~+ Reasoner}}} & - & {\textbf{53.60} \color[HTML]{3635F5}  (+1.95)} & {\textbf{43.42} \color[HTML]{3635F5} (+4.96)}& {\textbf{26.36} \color[HTML]{3635F5}  (+1.51)} & {\textbf{74.99} \color[HTML]{3635F5}  (+7.94)}  & {\textbf{57.78} \color[HTML]{3635F5}  (+18.94)}  \\
\midrule
\textbf{InstructBLIP}
& \textbf{13B} & 52.72 & 37.61 & 22.21& 58.38  & 23.46   \\
\rowcolor[HTML]{EFEFEF} 
\multicolumn{1}{l}{\cellcolor[HTML]{EFEFEF}{\textbf{~~+ Tools}}} & - & {51.92 \color[HTML]{cb0000} (-0.80)} & {\textbf{39.65} \color[HTML]{3635F5}  (+2.04)} & - & {69.29 \color[HTML]{3635F5} (+10.91)} & {51.37 \color[HTML]{3635F5} (+27.91)}  \\
\rowcolor[HTML]{EFEFEF} 
\multicolumn{1}{l}{\cellcolor[HTML]{EFEFEF}{\textbf{~~+ Reasoner}}} & - & {\textbf{54.24} \color[HTML]{3635F5} (+1.48)} & {\textbf{42.03} \color[HTML]{3635F5}  (+4.42)} & {\textbf{23.72} \color[HTML]{3635F5} (+1.51)}& {\textbf{74.98} \color[HTML]{3635F5} (+16.60)} & {\textbf{59.73} \color[HTML]{3635F5} (+36.27)}  \\
\midrule

\textbf{LLaVA}
& \textbf{13B} & 57.02 & 57.04 & - & 70.29  & 29.69      \\
\rowcolor[HTML]{EFEFEF} 
\multicolumn{1}{l}{\cellcolor[HTML]{EFEFEF}\textbf{{~~+ Tools}}} & - & {54.19 \color[HTML]{cb0000} (-2.83)} & {58.91 \color[HTML]{3635F5} (+1.87)} & - & {77.01 \color[HTML]{3635F5}  (+6.72)}  & {61.67 \color[HTML]{3635F5} (+31.98)}     \\ 
\rowcolor[HTML]{EFEFEF} 
\multicolumn{1}{l}{\cellcolor[HTML]{EFEFEF}\textbf{{~~+ Reasoner}}} & - & {\textbf{60.65} \color[HTML]{3635F5} (+3.63)} & {\textbf{59.22} \color[HTML]{3635F5} (+2.18)} & - & {\textbf{80.28} \color[HTML]{3635F5}  (+9.99)}  & {\textbf{68.65} \color[HTML]{3635F5} (+38.96)}     \\ 
\bottomrule
\end{tabular}
\caption{Evaluation results with different VLMs as the \textsf{Answer} tools. \textbf{Size} refers to the number of parameters of the language model in each VLM. We do not provide the result of LLaVA on ST-VQA because its output always includes abundant information, leading to nonsensical low scores using the official evaluation website. We highlight the best results in \textbf{bold}.}
\label{table: table1}
\end{table*}

\subsection{Baselines}
To comprehensively compare our approach with alternative task decomposition strategies, we devise a baseline termed ``Tools'' inspired by \citet{gupta2023visual}. 
Specifically, we employ a few-shot approach to decompose the initial problem into a sequence of tool-calling operations (e.g., PaddleOCR and GroundingDINO), which are subsequently executed. Notably, we leverage LLaMA-3-8B-Instruct~\cite{dubey2024llama} for decomposition, ensuring a comparable parameter count to the LLaVA-7B model used by our Reasoner. 
\subsection{Main Results}
Table~\ref{table: table1} reports evaluation results. 
We find that
\textbf{(1) The Reasoner consistently improves the performance of all VLMs across all datasets.}
By decomposing questions and invoking specialized tools, the Reasoner can improve various off-the-shelf VLMs in a plug-and-play fashion, suggesting its strong generalization ability.
\textbf{(2) The Reasoner helps better capture complex inter-entity relations.} The Reasoner brings clear improvements on GQA, which involves diverse relations among entities in the images. The improvements could result from the least-to-most synthesis algorithm in which such relations are implicitly modeled. The improvement for LLaVA~(3.63\%) is more substantial than other weaker VLMs~(1.17\%-1.95\%). It is possibly because stronger VLMs may better realize their potential with the Reasoner considerably alleviating the difficulty in locating the target entity.
\textbf{(3) The Reasoner boosts the understanding of words in images.} The \textsf{OCR} tool empowers the Reasoner to effectively enhance the performance on both TextVQA and ST-VQA. On TextVQA, we notice that the enhancement is less significant on LLaVA than on weaker VLMs, possibly because LLaVA is less dependent on external tools for understanding the words in images. 
\textbf{(4) The Reasoner improves the counting performance by a large margin.} 
The benefit of the Reasoner is particularly significant on TallyQA.
Without the Reasoner, the base VLMs are sensitive to irrelevant information in the images and thus easily miscount. By utilizing the \textsf{Highlight} tool, the Reasoner can reliably mitigate the impact of such irrelevant information.

\subsection{Quality Assessment of \textsc{Vireo}}
We conduct a human study to investigate the quality of our synthesized \textsc{Vireo} dataset. Specifically, we randomly sample $200$ examples from \textsc{Vireo}, and recruit $3$ graduate students majoring in NLP to manually check the correctness of the synthesized question and reasoning process for each instance. 
The check focuses on the correctness in three aspects: \textbf{(1) Sub-Question.} The sub-question must be accurately faithful to the profile of the tail node and can be resolved using the specified tool. \textbf{(2) Argument.} Based on the current sub-question, the annotators need to assess whether using the given argument to invoke the tool can yield the expected output. \textbf{(3) Main Question.} The annotators judge whether the main question covers all sub-questions and whether it can be decomposed into sub-questions in the correct order. Each example receives 3 labels from each of the three annotators on each aspect. 
Results show that the three evaluators consistently approve that all examples are correct in all aspects\footnote{This might be surprising, but we double-checked the evaluation and confirmed the results.} 
, suggesting the highly guaranteed quality of the synthesized data. 

\section{Discussions}
Although the main results affirmed the high quality of \textsc{Vireo} and the general benefits of the Reasoner to enhance VLMs on multiple benchmarks, we are still curious about the following research questions:
\textbf{(1) RQ1: }How does the capability of the Reasoner vary w.r.t. the size of instructions for fine-tuning?
\textbf{(2) RQ2:} 
What is the relative impact of each integrated tool on the Reasoner's overall capability? 
 \textbf{(3) RQ3:} Will the Reasoner bring a negative impact on visual tasks other than VQA?
\textbf{(4) RQ4: } Is the Reasoner still effective on more advanced VLMs?
\textbf{(5) RQ5:} What are the underlying causes of the Reasoner's errors on the benchmarks?

\subsection{RQ1: Influence of Data Size}

\begin{figure}[!t]
    \centering
    \includegraphics[width=0.48\textwidth]{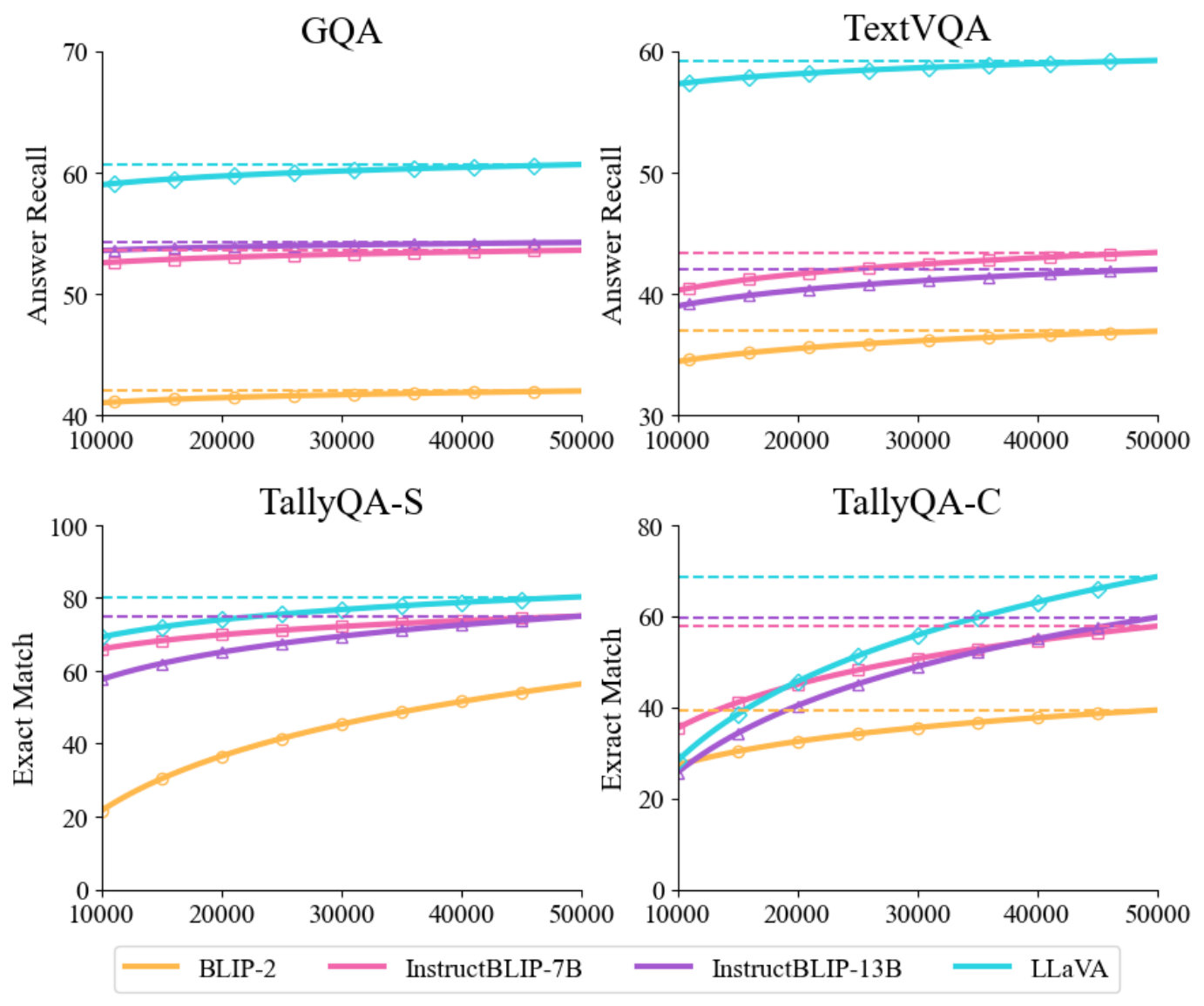}
    \caption{The performance of the Reasoner varies with the size of \textsc{Vireo}. The dashed lines indicate the performance of corresponding models using 50k training examples.}
    \label{fig:scaling}
\end{figure}

\begin{table}[!tbp]
\centering
\resizebox{0.95\columnwidth}{!}{\begin{tabular}{lcccc}
\toprule
\multicolumn{1}{l}{\textbf{Model}} & \multicolumn{1}{l}{\textbf{GQA}} & \multicolumn{1}{l}{\textbf{TextVQA}} & \multicolumn{1}{l}{\textbf{Tally-S}} & \multicolumn{1}{l}{\textbf{Tally-C}} \\ \midrule
\rowcolor[HTML]{EFEFEF} 
\textbf{BLIP2} & \textbf{42.02} & \textbf{36.94} & \textbf{56.38} & \textbf{39.38} \\
w/o OCR & 40.26 & 32.30 & 54.89 & 37.89 \\
w/o Highlight & 39.34 & 36.30 & 20.45 & 28.66 \\
w/o Grounding & 40.44 & 31.98 & 20.89 & 28.84 \\ \midrule
\rowcolor[HTML]{EFEFEF} 
\textbf{InstBLIP-7B} & \textbf{53.60} & \textbf{43.42} & \textbf{74.99} & \textbf{57.78} \\
w/o OCR & 50.09 & 37.20 & 73.17 & 55.78 \\
w/o Highlight & 48.90 & 41.78 & 65.68 & 38.49 \\
w/o Grounding & 51.03 & 37.84 & 67.01 & 38.67 \\ \midrule
\rowcolor[HTML]{EFEFEF} 
\textbf{InstBLIP-13B} & \textbf{54.24} & \textbf{42.03} & \textbf{74.98} & \textbf{59.73} \\
w/o OCR & 50.25 & 36.56 & 73.19 & 55.78 \\
w/o Highlight & 48.90 & 41.70 & 57.51 & 25.11 \\
w/o Grounding & 51.82 & 37.46 & 58.61 & 25.14 \\ \midrule
\rowcolor[HTML]{EFEFEF} 
\textbf{LLaVA} & \textbf{60.65} & \textbf{59.22} & \textbf{80.28} & \textbf{68.65} \\
w/o OCR & 53.86 & 55.89 & 77.28 & 64.53 \\
w/o Highlight & 51.82 & 37.46 & 58.61 & 25.14 \\
w/o Grounding & 56.51 & 58.09 & 70.22 & 33.04 \\ \bottomrule
\end{tabular}}
\caption{Evaluation results of removing different tools.}
\label{table:discussion_table2}
\end{table}

Figure~\ref{fig:scaling} presents the model performance when we vary the size of \textsc{Vireo}. Overall, the gain in performance gradually increases with the number of training examples. However, we also notice that the marginal benefit is diminishing,
suggesting that more diverse data synthesis approaches besides ours are needed to further break through the upper limit in future work.



\subsection{RQ2: Ablation Studies Regarding Tools}

To gauge the significance of the diverse tools employed in our approach, we conducted a thorough ablation study, systematically investigating the impact of removing each tool on model performance. The findings, presented in Table~\ref{table:discussion_table2}, unequivocally demonstrate that the exclusion of any single tool adversely affects the model's capabilities across various datasets. This observation underscores the multifaceted nature of visual tasks, which often necessitate the synergistic application of multiple sophisticated operations, rendering the reliance on a solitary tool insufficient.

\subsection{RQ3: Extending to Broader Visual Tasks}
To investigate whether introducing a Reasoner will hurt the performance on other visual tasks, 
we conduct experiments on MMMU~\cite{yue2024mmmu} and POPE~\cite{li2023evaluating} as examples. The MMMU benchmark aggregates massive multi-discipline tasks necessitating abundant knowledge to address, which is beyond the scope of our Reasoner.
POPE is a 
discrimination dataset for hallucination detection 
which 
requires discriminating whether a certain coarse-grained object is present in a given image without the need for multi-step reasoning. We choose InstructBLIP-7B and -13B as the base VLMs for discussion.


\begin{table}[!tbp]
\centering
\begin{adjustbox}{max width=\linewidth}

\begin{tabular}{lcll}
\toprule
\textbf{Model}  & \textbf{Size}        & \textbf{MMMU} & \textbf{POPE} \\ \midrule
\textbf{InstructBLIP} & \textbf{7B} & 22.43         & 84.79         \\
\rowcolor[HTML]{EFEFEF} 
\textbf{~~+ Reasoner}                     & -               & 22.50 \color[HTML]{3635F5} (+0.07)         & 84.79  \color[HTML]{3635F5} (+0.00)       \\ 
                     
                     \midrule
\textbf{InstructBLIP} & \textbf{13B} & 18.75         & 80.42         \\
\rowcolor[HTML]{EFEFEF} 
\textbf{~~+ Reasoner} & - & 18.23 \color[HTML]{3635F5} (-0.52)               & 80.42 \color[HTML]{3635F5} (+0.00)         \\ \bottomrule
\end{tabular}
\end{adjustbox}
\caption{Results on broader visual tasks beyond VQA.}
\label{table:discussion_table1}
\end{table}

As shown in Table~\ref{table:discussion_table1}, introducing the Reasoner on MMMU has a minimal impact on the final performance, 
suggesting that the Reasoner does not influence the expression of VLMs' inherent knowledge. 
On the other hand, the presence or absence of the Reasoner on POPE yields consistent results, indicating that additional reasoning over images does not increase the likelihood of the VLMs generating hallucinations.


\subsection{RQ4: Efficacy on More Advanced VLMs}
Considering that more advanced VLMs such as Qwen-VL~\cite{bai2023qwen} have huge potential to inherit the function of external tools such as perceiving bounding boxes in images, we convert \textsc{Vireo} into an end-to-end format by sequentially combining the input and output of each step in order to avoid the computation overhead from explicit tool invocation.
Table~\ref{appendix: qwen} shows that the Qwen-VL model, fine-tuned on the end-to-end version of \textsc{Vireo}, significantly improves across all datasets. Notably, the application of our Reasoner does not cause a noticeable increase in time cost compared to the vanilla model (1.4s v.s. 1.1s per sample).

\begin{table}[!tbp]
\centering
\resizebox{\columnwidth}{!}{\begin{tabular}{lcccc}
\toprule
\textbf{Model} & \multicolumn{1}{l}{\textbf{GQA}} & \multicolumn{1}{l}{\textbf{TextVQA}} & \multicolumn{1}{l}{\textbf{TallyQA-S}} & \multicolumn{1}{l}{\textbf{TallyQA-C}} \\ \midrule
\textbf{Qwen-VL} & 57.95 & 67.26 & 75.38 & 37.92 \\
\rowcolor[HTML]{EFEFEF} 
\multicolumn{1}{r}{\cellcolor[HTML]{EFEFEF}\textbf{~~+ Reasoner}} & 61.73 & 74.12 & 81.50 & 68.40 \\ \bottomrule
\end{tabular}}
\caption{Evaluation results on Qwen-VL.}
\label{appendix: qwen}
\end{table}

\subsection{RQ5: Regarding Error Analysis}
To analyze the error types of the Reasoner, we randomly sample 100 instances where the model prediction is wrong from {all evaluation sets}. Then, we analyze their error 
and categorize the errors into three types: \textbf{(1) Reasoning:} The Reasoner uses a wrong tool~(\textit{Tool}) or generates wrong arguments for the tool~(\textit{Arguments}); \textbf{(2) Execution:} The \textit{\textsf{Grounding}}, \textit{\textsf{OCR}}, or \textit{\textsf{Highlight}} tool returns wrong execution results; and \textbf{(3) Inference:} The \textsf{Answer} tool outputs wrong answers~(\textit{Wrong}) or irrelevant answers with the question~(\textit{Missing}).
\begin{figure}[!t]
    \centering
    \includegraphics[width=0.35\textwidth]{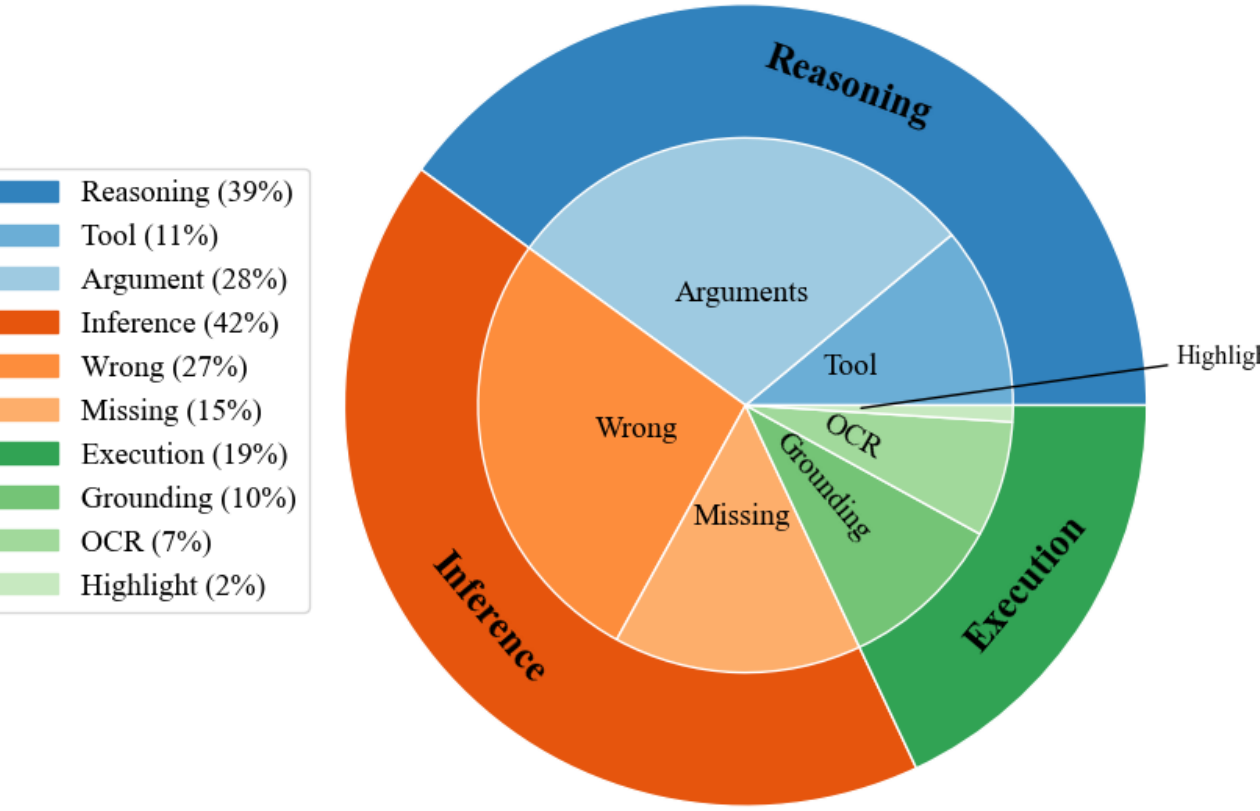}
    \caption{The distribution of different error types.}
    \label{fig:error_analysis}
\end{figure}
\textcolor{red}{
}

As shown in Figure~\ref{fig:error_analysis}, the ``Inference'' error has the largest proportion~(42\%). Among these, 
there are even 15\% of instances where the \textsf{Answer} tool generates irrelevant answers, indicating the huge room to improve the instruction-following ability of existing VLMs. 
Additionally, the "Reasoning" error accounts for a significant proportion (39\%), and most of the instances~(28\%) are attributed to wrong arguments. This means VLMs' tool-invoking capability is far from perfect, yet this ability is crucial for VLMs to interact with the environment.
The ``Execution'' error is less frequent, implying the huge potential of building general and powerful VLMs based on specialized visual tools.





\section{Conclusions}
We propose a data synthesis approach to enhancing multi-step reasoning capabilities of vision-language models. The approach decomposes the synthesis task into several simple sub-tasks, and finish the sub-tasks (almost) with open-sourced models. Based on the approach, we build a large-scale visual reasoning dataset, and develop a visual reasoner by tuning on the dataset. Evaluation results across four VQA benchmarks indicate that the visual reasoner can generally improve the reasoning capabilities of a number of existing VLMs.

\section*{Limitations}
We select COCO2014~\cite{lin2014microsoft} as the source of images for our data synthetic process. However, while COCO2014 is a general-purpose image dataset, we do not guarantee that its data can cover all visual tasks. 
Additionally, our proposed method demonstrates consistent improvements across four VQA benchmarks, but this does not imply that our method will be effective in all visual datasets and scenarios. Furthermore, during our experiments, we only selected four VLMs as base models, and we cannot ensure that our method is capable of bringing improvements to all models.

\section*{Ethical Considerations}
Images may contain sensitive information. Using or publishing datasets that include such information could pose potential ethical risks. In our experiments, we strictly control the image sources of our data, utilizing only authorized open-source datasets. Furthermore, all the models we used are publicly accessible and adhere to established requirements.

\section*{Acknowledgments}
This work was supported by the National Natural Science Foundation of China (NSFC Grant No. 62122089), Beijing Outstanding Young Scientist Program NO. BJJWZYJH012019100020098, and Intelligent Social Governance Platform, Major Innovation \& Planning Interdisciplinary Platform for the ``Double-First Class'' Initiative, Renmin University of China, the Fundamental Research Funds for the Central Universities, and the Research Funds of Renmin University of China.



\appendix


\label{sec:details}
\section{Details of Data Construction.}
\label{appendix:construction}

\paragraph{Attributes of Single-Entity Node.}
Each single-entity node consists of five attributes: \textit{Label, Location, Color, Text, and Size.} Specifically, 
\textit{Label} and \textit{Location} are directly derived from the recognition results of Deformable DETR. 
\textit{Color} is determined using ColorThief~\footnote{\url{https://lokeshdhakar.com/projects/color-thief/}} to analyze the dominant color scheme of the subgraph corresponding to the single-entity node. 
\textit{Text} is obtained by applying PaddleOCR to recognize textual content within the image. 
\textit{Size} records the node's proportion in the entire image, considering its width, height, and area.

\paragraph{Construction of Chain.}
We use tools that produce graphical outputs (e.g., Grounding, Highlight) to connect intermediary nodes and tools that produce textual outputs (e.g., OCR, Answer) to connect terminal nodes. 

We construct the chain by sequentially adding nodes. Specifically, we maintain a queue \( L = (N_1, N_2, \ldots, N_i) \) consisting of nodes. To add \( N_{i+1} \), we first specify the tool \( t_i \) to be used and sample a node in the remaining set \( \hat{N} = \{ N_j \notin L \} \) that allow the use of this tool. For instance, only nodes with text in images permit using the OCR tool. If \( \hat{N} = \emptyset \) or the chain length reaches the limit, the process terminates. 
When constructing \textsc{Vireo}, we set the maximum chain length to 4. Consequently, the dataset includes reasoning paths with lengths of 2, 3, and 4.

\section{Implementation Details}
\label{appendix:experiment}

\paragraph{Image Source.}
We use the COCO2014 dataset~\cite{lin2014microsoft} as our source of images.

\paragraph{Questioner.}
We use LLaMA-3-8B-Instruct as the base model for the \texttt{Questioner}. 
To construct the corresponding training data, we first manually create 5 seed prompts for each combination of head node and tail node ($3 \times 3 = 9$). Then, we generate a total of 10k instances by using GPT-4. 
During this process, to reduce the bias introduced by the seed prompts, we gradually add the obtained results into the prompt pool. For each instance, three prompts are randomly selected from the prompt pool as demonstrations. 
For each instance, the input contains 3 fields: the profiles of the head node and tail node, and the specified tool. Output includes 2 fields: the question and the arguments of the tool.
We perform instruction fine-tuning on the Questioner using LoRA.
The rank is set to 8 and the lora\_alpha is set to 8. We adopt AdamW as the optimizer. We set adam\_beta1, adam\_beta2, and adam\_epslion to 0.9, 0.999, and 1e-8, respectively. We use the cosine schedule to warm up the training and set warmup\_steps to 0.1. We set the batch size to 4 and fine-tune Questioner for 2 epochs, with each epoch taking around 1 hour. The training process is completed with 8 Nvidia A100 GPUs.

\paragraph{Combiner.}
Similar to \texttt{Questioner}, we also use LLaMA-3-8B-Instruct as the base model for the \texttt{Combiner}. We manually create 20 seed prompts and generate a total of 10k instances by using GPT-4. The input includes two questions to be merged, and the output is the merged result. The fine-tuning settings of \texttt{Combiner} keep the same as \texttt{Questioner}.

\begin{table}
\begin{tabular}{lccc}
\toprule
\textbf{Model} & \multicolumn{1}{l}{\textbf{Size}} & \multicolumn{1}{l}{\textbf{VLM-only}} & \multicolumn{1}{l}{\textbf{Reasoner}} \\ \midrule
BLIP2 & 2.7B & 0.05 & 1.40 \\
\rowcolor[HTML]{EFEFEF} 
InstructBLIP & 7B & 0.19 & 1.70 \\
InstructBLIP & 13B & 0.28 & 1.79 \\
\rowcolor[HTML]{EFEFEF} 
LLaVA & 13B & 0.57 & 2.08 \\ \bottomrule
\end{tabular}
\caption{The average per-sample time cost (seconds) when using only the VLM and applying the Reasoner.}
\label{table:discussion_table3}
\end{table}

\paragraph{Reasoner.}
We synthesize 50k data examples using the least-to-most synthesis method. Each example includes the current image \( I \), the main question \( Q \), and the previous sub-questions \( \{q_{<k}\} \). The label for each data example comprises the current sub-question \( q_k \) and the tool \( t_k \) that needs to be invoked. 
We use this data to train LLaVA-1.5-7B as the Reasoner. We perform instruction fine-tuning on the Reasoner using LoRA. The rank is set to 8 and the lora\_alpha is set to 8. We adopt AdamW as the optimizer. We set adam\_beta1, adam\_beta2, and adam\_epslion to 0.9, 0.999, and 1e-8, respectively. We use the cosine schedule to warm up the training and set warmup\_steps to 0.1. We set the batch size to 8 and fine-tune the Reasoner for 3 epochs, with each epoch taking around 3 hours. The training process is completed with 1 Nvidia A100 GPU.


\section{Additional Time Cost}
While the Visual Reasoner offers significant performance gains across various tasks, it is crucial to acknowledge the computational overhead associated with its multi-step reasoning process. As each step necessitates additional inference time, the overall inference duration is extended compared to using a vanilla VLM. Table~\ref{table:discussion_table3} presents a quantitative analysis of the average per-sample time cost using a single Nvidia A100 GPU.

\section{Case Study}
\begin{figure*}[htbp]
    \centering
    \includegraphics[width=0.9\textwidth]{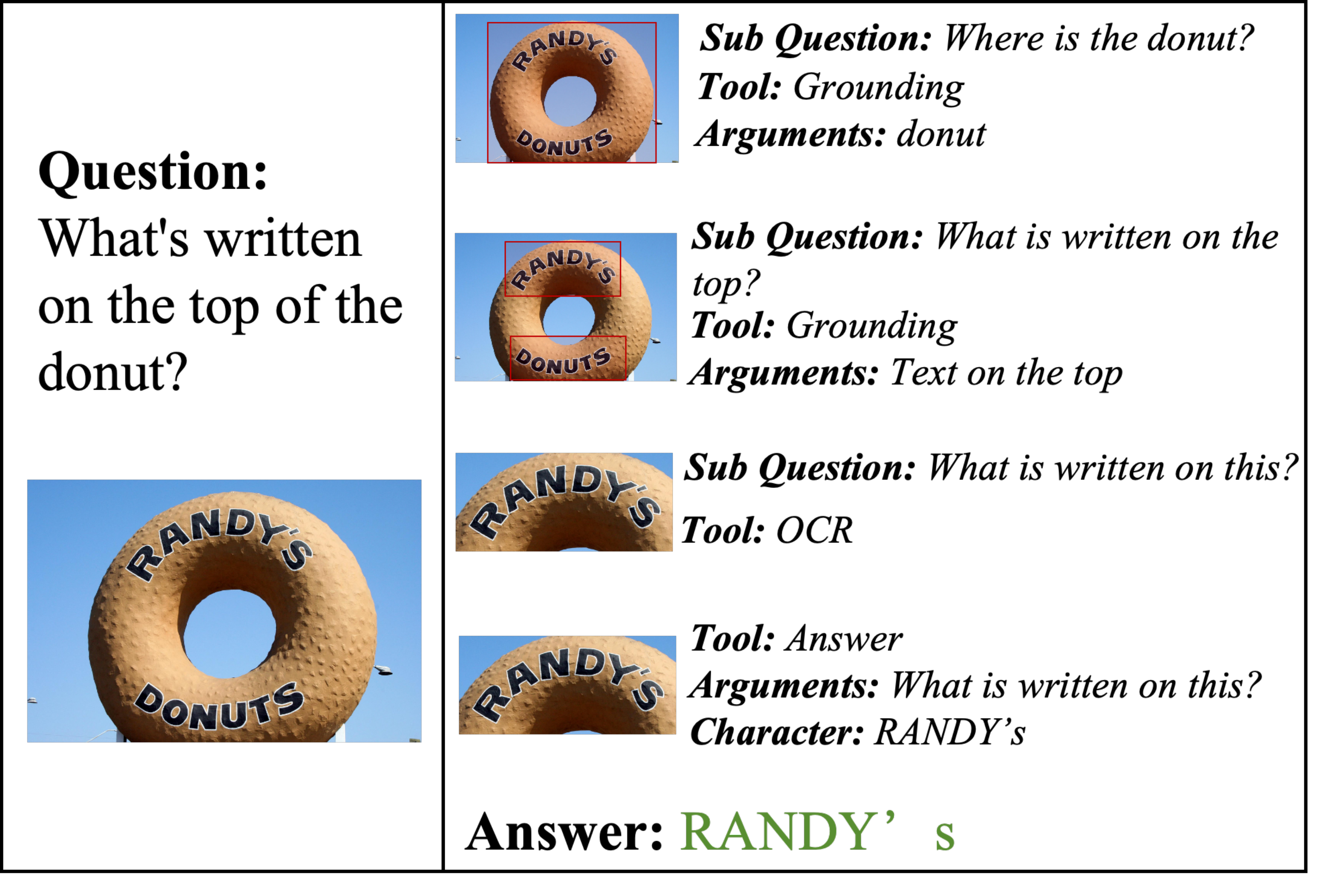}
    \caption{Case 1. This case uses Grounding to locate the donut and uses OCR and Answer to get the final answer.}
    \label{fig:case_study1}
\end{figure*}

\begin{figure*}[htbp]
    \centering
    \includegraphics[width=0.9\textwidth]{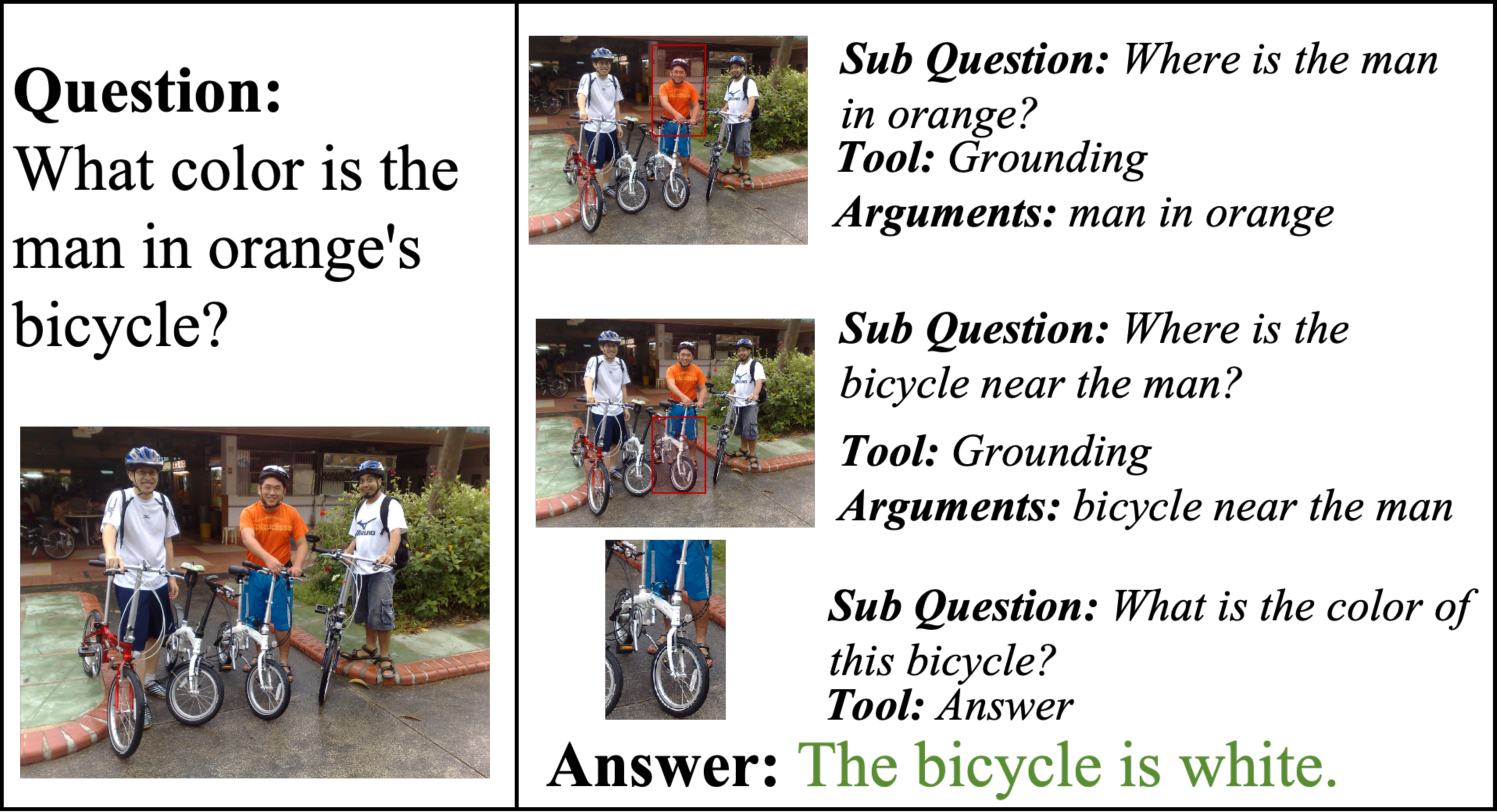}
    \caption{Case 2. In this case, the man wearing orange is first precisely identified, and then attention is directed to the bicycle near him to obtain the answer.}
    \label{fig:case_study2}
\end{figure*}

\begin{figure*}[htbp]
    \centering
    \includegraphics[width=0.9\textwidth]{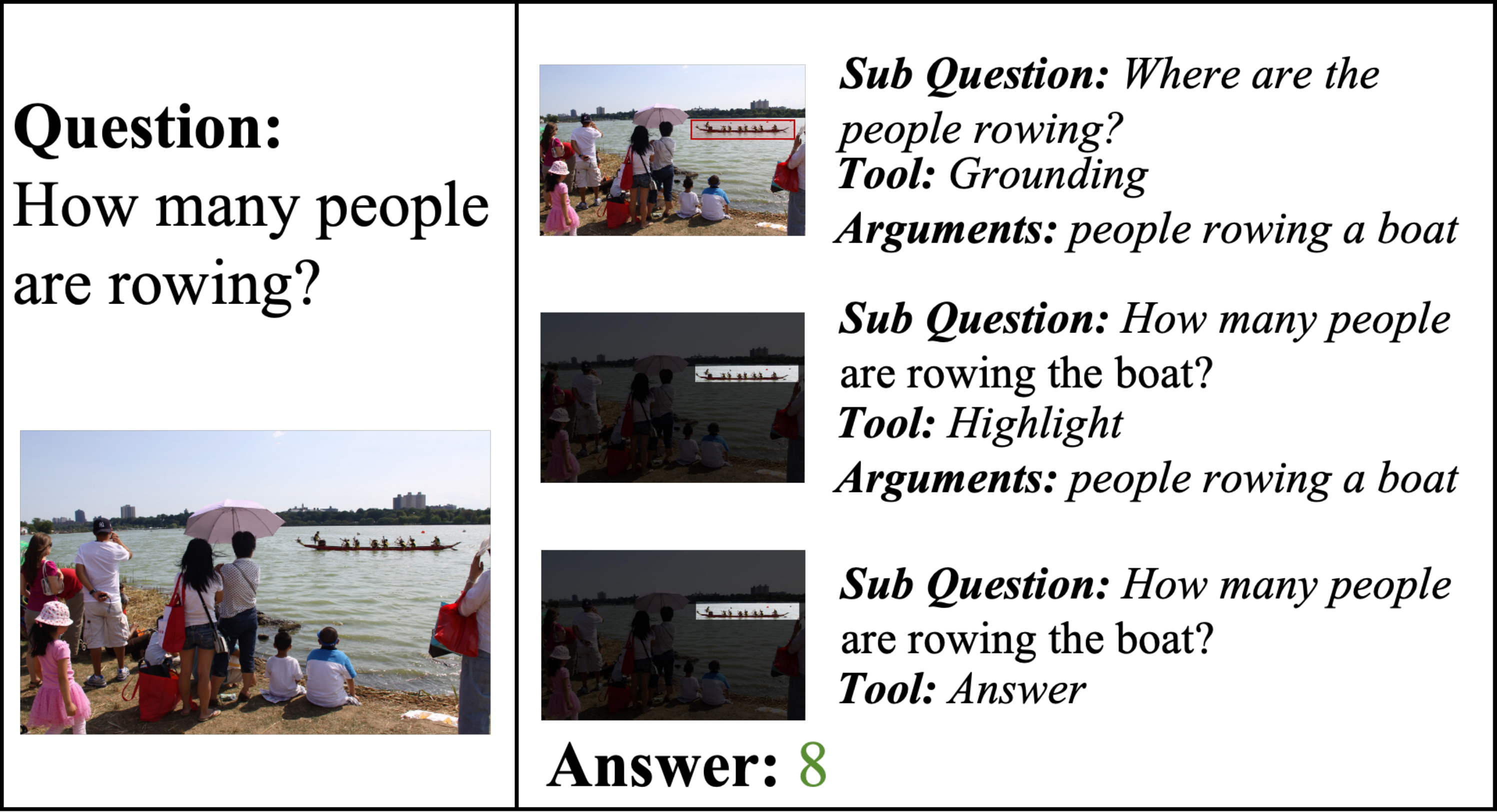}
    \caption{Case 3. This case excludes irrelevant people and highlights the target group, thereby achieving accurate counting results.}
    \label{fig:case_study3}
\end{figure*}

To facilitate understanding of our method, we show some cases in Figure \ref{fig:case_study1}, Figure \ref{fig:case_study2}, and Figure \ref{fig:case_study3}.

\section{Example of Data Construction}
\begin{figure*}[htbp]
    \centering
    \includegraphics[width=0.7\linewidth]{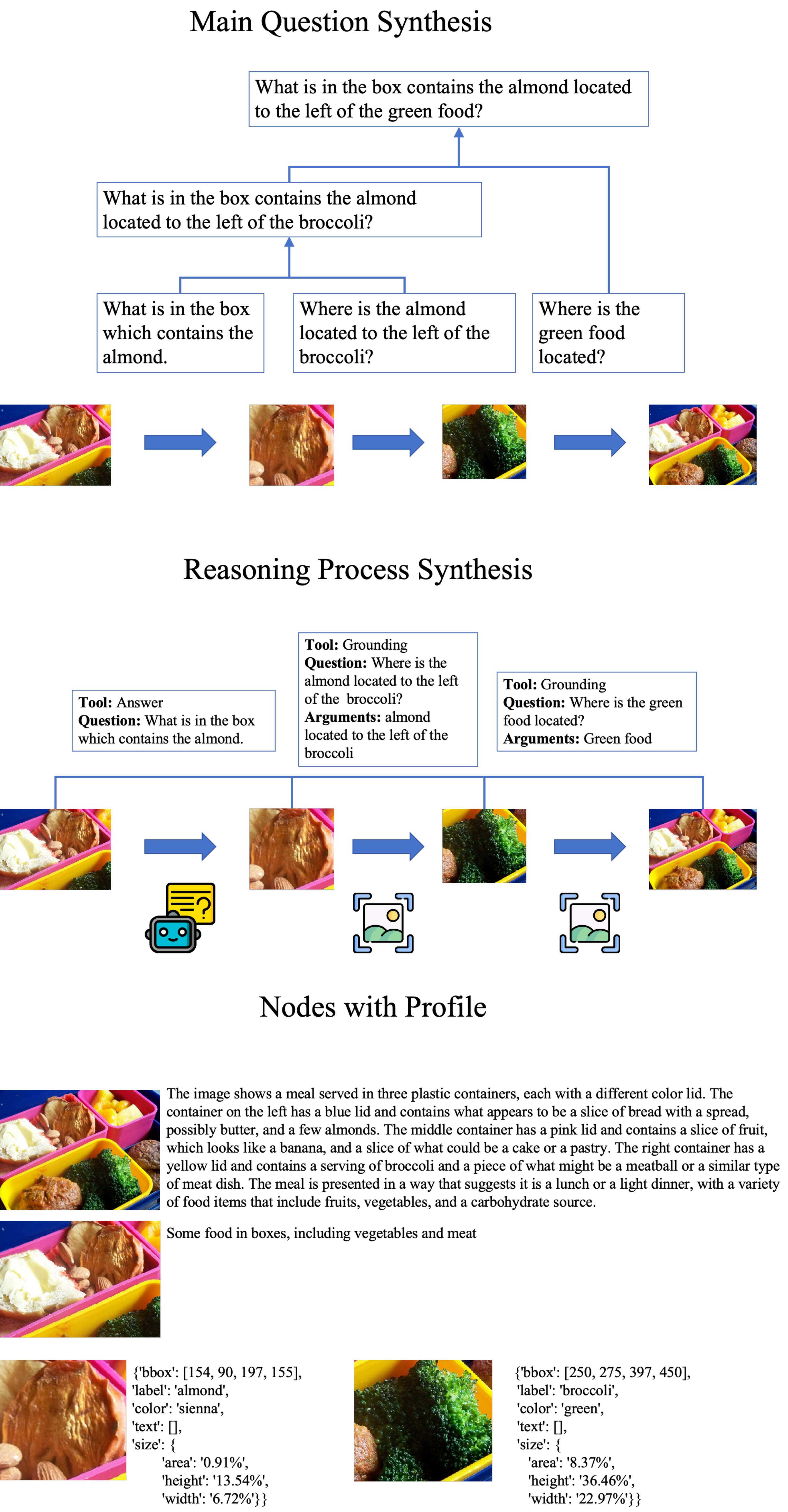}
    \caption{A simple case to demonstrate the data construction process.}
    \label{fig:synthesis}
\end{figure*}

To visualize our least-to-most pipeline, we show the data construction process in Figure~\ref{fig:synthesis}.

\end{document}